\documentclass{article}
\usepackage{lmodern}
\usepackage[a4paper, total={6in,9.75in}]{geometry}
\usepackage{graphicx} 
\usepackage{authblk}
\usepackage{subcaption} 
\usepackage{wrapfig}
\usepackage[most]{tcolorbox}

\usepackage[table]{xcolor}
\usepackage{colortbl}
\usepackage{booktabs}
\usepackage{mathtools}
\usepackage{amsmath, amssymb, amsfonts}
\usepackage{hyperref}
\usepackage{tabularx}
\usepackage{array} 
\usepackage{float}
\usepackage{multirow}
\usepackage{tikz}
\usetikzlibrary{positioning, matrix, arrows, calc, arrows.meta, shapes, fit}

\usepackage[skip=4pt]{caption}

\usepackage{titlesec}
\titlespacing{\section}{0pt}{6pt}{4pt}
\titlespacing{\subsection}{0pt}{4pt}{2pt}

\usepackage{enumitem}
\setlist{nosep}

\usepackage{makecell}

\usepackage{amssymb}
\newcommand{\cmark}{$\checkmark$}
\newcommand{\xmark}{$\times$}
\newcommand{\pmark}{$\sim$}

\usepackage{comment}
\usepackage{etoolbox} 
\DeclareGraphicsExtensions{.pdf,.png,.jpg}

\author{Adeela Bashir$^1$}
\author{Zia Ush Shamszaman$^{1,2}$}
\author{Zhao Song$^1$}
\author{Matjaz Perc$^{3,4,5,6,*}$}
\author{The Anh Han$^{1,2,*}$}
\affil{$^{1}$School of Computing, Engineering and Digital Technologies, Teesside University\\ $^{2}$Center for Digital Innovation, Teesside University\\ 
 $^3$ Faculty of Natural Sciences and Mathematics, University of Maribor, Maribor, Slovenia\\
  $^4$ Community Healthcare Center Dr. Adolf Drolc Maribor, Maribor, Slovenia\\
  $^5$ University College, Korea University, Seoul, Republic of Korea\\
  $^6$ Department of Physics, Kyung Hee University, Seoul, Republic of Korea\\ 
$^*$Corresponding: The Anh Han (T.Han@tees.ac.uk), Matjaz Perc (matjaz.perc@gmail.com)}

\begin{document}

\title{Strategic commitments shape collective cybersecurity under AI inequality} 

\date{}
\maketitle

\begin{abstract}
The growing integration of AI into cybersecurity is reshaping the balance between attackers and defenders. When access to advanced AI-enabled defence tools is uneven, resource-limited defenders may be unable to adopt effective protection, creating persistent system vulnerabilities. We study the impact of differential AI access using an evolutionary game-theoretic model in a finite population. We first show that when high-capability defence is costly, the population is driven toward low-cost, weak-defence behaviour, sustaining attacks and weakening long-run security. To address this problem, we introduce differential access to AI defence tools by allowing defenders to choose between low- and high-capability protection based on their resources. We then examine the role of a small group of committed defenders who always adopt strong defence and influence others through social learning. Although commitment increases the prevalence of strong defence, it alone cannot stabilise secure outcomes due to high defence costs. We therefore incorporate a targeted subsidy to remove the cost disadvantage from committed defenders. Our analysis shows that subsidised commitment significantly increases strong defence adoption, suppresses successful attacks, and improves overall system resilience. Simulations across a broad parameter space confirm that subsidies consistently outperform commitment alone. In addition, social-welfare analysis shows improved defender outcomes while keeping attacker gains low. These findings suggest that targeted support for key defenders can be an effective mechanism for stabilising cybersecurity in AI-driven environments and provide a theoretical bridge between cybersecurity policy, AI governance, and strategic allocation of defensive AI capabilities.
\end{abstract}

\noindent\textbf{Keywords: }{\textit{Evolutionary Game Theory; Cyber Security; AI Security; Finite Population Dynamics; Differential AI Access; Committed Defenders; Social Welfare; Attack–Defence Strategies}

\section{Introduction}
\label{1}

Artificial intelligence (AI) is rapidly transforming cyber offence and defence \cite{heinl2014artificial,truong2020artificial}. AI-powered security systems improve anomaly detection, automated response, and threat prediction, while attackers deploy the same technologies to scale intrusion, craft adaptive phishing, and generate adversarial attacks \cite{chakraborty2021survey,ajala2024leveraging}. As a result, the frequency and sophistication of AI-enabled cyberattacks continue to rise, impacting organisations with limited defensive resources \cite{zdrojewski2025ai}. 

A growing challenge is the unequal access to advanced defence capabilities. Small and medium enterprises (SMEs), which face nearly half of global cyberattacks \cite{tetteh2024cybersecurity}, often cannot afford costly security infrastructure, specialised staff, or advanced AI defence tools. This resource asymmetry results in persistent vulnerability and forces many defenders to adopt weaker, low-cost protection strategies. Recent study on AI safety refer to this imbalance as \textit{differential access} to AI security tools \cite{ee2025asymmetry}, highlighting how unequal distribution of defensive capability can shape systemic cyber risk. To study this phenomenon analytically, we model two classes of defenders: $H$ defenders using advanced, costly AI defence tools, and $L$ defenders using basic tools. We analyse strategic interaction between attackers and defenders in a finite population using stochastic EGT, allowing realistic modelling of population dynamics, fixation, and transition probabilities \cite{traulsen2006stochastic,masuda2012evolution}. Although H-defence is more effective, its high cost drives many defenders toward L-defence, resulting in attack success and reducing overall social welfare. To encourage adoption of strong defence, we introduce a subset of \emph{committed defenders} $(z)$ who always choose H-defence—representing government agencies, cybersecurity vendors, or regulated institutions capable of absorbing high costs. These players act as persistent examples of secure behaviour which drive the population's behavioural dynamics through social learning \cite{mahmud2026eco,nakajima2015evolutionary}. Results show that committed defenders improve the frequency of H-defence but do not improve social welfare due to high defence cost and persistent attack incentives. We therefore propose a policy mechanism: subsidising the cost of committed defenders (e.g., government funding), aiming to enable rapid growth of high-defence behaviour and increase collective security without requiring universal cost reduction.
\textit{We study under what parameter conditions and policy interventions can committed defenders shift a competitive attacker–defender ecosystem toward a stable high-defence, low-attack equilibrium and improved social welfare?}

Our results highlight that strengthening defence incentives—rather than penalising attackers—is a more effective mechanism for stabilising cyber ecosystems. In particular, targeted support for a small subset of trusted defenders can trigger system-wide improvements in security and resilience. These findings provide a theoretical foundation for differential AI access policies, including selective deployment of advanced defence tools, certification of trusted defenders, and targeted subsidies for critical infrastructure protection. Such approaches enable economically efficient cybersecurity without requiring universal adoption of costly defence technologies.

In the rest of this paper, Section \ref{2} reviews recent advances in cyber defence, AI-driven security, game-theoretic modelling and compare our work with prior research in this domain. Section \ref{3} introduces our finite-population game-theoretic framework for attacker–defender interactions. Section \ref{4} presents the concept of strategic AI access allocation and analyses population dynamics using stochastic evolutionary processes. Section \ref{5} introduces a subsidy mechanism for committed defenders and demonstrates its impact on system-wide resilience and evaluates collective safety through a social welfare analysis. Section \ref{7} discusses the implications of our findings, connects them to emerging AI governance practices and outlines key research directions. Finally, Section \ref{8} concludes the work by summarising the core theoretical and policy contributions.

\section{Related Work}
\label{2}

Cyber defence has increasingly adopted game-theoretic modelling to analyse strategic behaviour between attackers and defenders under uncertainty. EGT offers a powerful framework for studying cyber conflict and strategic interactions in dynamic populations, particularly where learning, imitation, and bounded rationality drive strategy evolution \cite{bashir2026co,bashir2026trust,d2015statistical,perc2017statistical,SONG2026116603,QU2026118214}.
Recent studies demonstrate the value of stochastic finite-population analysis over traditional infinite-population and deterministic models, enabling close predictions of human experimental data \cite{rand2013evolution,zisis2015generosity} and more realistic modelling of cyber environments such as industrial IoT, cloud systems, and autonomous defence ecosystems \cite{huang2015stochastic,bhat2025stochastic}.

Differential access to defence capability—where defenders vary in resources and expertise—has been highlighted as a critical challenge for cybersecurity resilience. Empirical studies report that SMEs face disproportionately high cyberattack rates but lack access to advanced security tools and skilled workforce \cite{junior2023unaware,perozzo2021assessing}. The AI governance community emphasises that unequal access to advanced AI defence systems may widen the security gap and create systemic risk \cite{taeihagh2025governance,papagiannidis2023toward}. However, formal analytical studies of unequal cyber–defence capability within strategic population models remain limited.

Committed or stubborn agents have been shown to shift equilibrium outcomes in social influence and cooperation models, enabling transitions to socially desirable states \cite{zimmaro2024emergence,hunter2022optimizing,cimpeanu2022artificial}. Their role in cybersecurity, particularly as a mechanism to promote robust defence behaviour in heterogeneous defender populations, remains largely unexplored. Moreover, existing works have studied subsidy and incentive mechanisms for cyber defence, showing that economic support can change defender behaviour and reduce attack success \cite{wang2019should,ge2024game}. However, how resource allocation interacts with evolutionary strategy dynamics and stochastic fixation in finite populations is still an open question. 

\begin{table}[h]
\centering
\renewcommand{\arraystretch}{1.2}
\caption{Comparative analysis of cyber security and evolutionary game-theoretic models.}
\label{tab:1}

\resizebox{1\linewidth}{!}{
\begin{tabular}{|c|p{6cm}|c|c|c|c|c|}
\hline
\textbf{Work} & \textbf{Model Type} 
& \makecell{\textbf{Finite} \\ \textbf{Population}} 
& \makecell{\textbf{Stochastic} \\ \textbf{Dynamics}} 
& \makecell{\textbf{Differential} \\ \textbf{Defence}} 
& \makecell{\textbf{Committed} \\ \textbf{Agents}} 
& \makecell{\textbf{Subsidy} \\ \textbf{Policy}} \\
\hline

\cite{ee2025asymmetry} 
& Conceptual AI governance model 
& \xmark 
& \xmark 
& \cmark 
& \pmark 
& \cmark \\
\hline

\cite{nakajima2015evolutionary} 
& Finite population evolutionary game 
& \cmark 
& \cmark 
& \xmark 
& \cmark 
& \xmark \\
\hline

\cite{bhat2025stochastic} 
& Stochastic field theory 
& \cmark 
& \cmark 
& \xmark 
& \xmark 
& \xmark \\
\hline

\cite{zimmaro2024emergence} 
& Evolutionary game with hybrid agents 
& \pmark 
& \cmark 
& \xmark 
& \cmark 
& \xmark \\
\hline

\cite{wang2019should} 
& Stackelberg security game 
& \xmark 
& \pmark 
& \pmark 
& \xmark 
& \cmark \\
\hline

\cite{ge2024game} 
& Multi-layer evolutionary game 
& \xmark 
& \pmark 
& \xmark 
& \xmark 
& \cmark \\
\hline

\textbf{This paper} 
& \textbf{Finite population stochastic EGT} 
& \textbf{\cmark} 
& \textbf{\cmark} 
& \textbf{\cmark} 
& \textbf{\cmark} 
& \textbf{\cmark} \\
\hline

\end{tabular}
}
\end{table}

While prior studies have examined aspects of cybersecurity or commitment using game-theoretic or evolutionary models, they typically consider these dimensions in isolation. In contrast, our work provides a unified framework where we model heterogeneous defenders with differential access to security resources, incorporate committed agents to capture persistent secure behaviour, and analyse the role of subsidy as a policy mechanism to improve system resilience. Table~\ref{tab:1} highlights the key distinctions between our work and existing literature. To the best of our knowledge, this is the first work that jointly studies stochastic population dynamics, differential defence, commitment, and subsidy-driven incentives in a cybersecurity context.

\section{Cybersecurity Game Theoretic Model}
\label{3}

We model cyber conflict as an asymmetric game between two interacting groups: attackers and defenders, within a well-mixed finite-population of size $N$. Attackers can choose among strategies attack $(A)$ and not attack $(NA)$, while defenders can choose strategies defence $(D)$ and not defence $(ND)$. We refer to each combination of attacker and defender strategies as a \textit{strategy pair}, for example $(A,D)$ or $(NA,ND)$. To analyse the strategic adaptation, we use EGT because it provides a suitable framework for behavioural evolution in cybersecurity \cite{bashir2026co,tosh2018establishing}. In addition, EGT assumes fully rational agents, captures bounded rationality and imitation-driven learning, where players adapt strategies based on observed payoff differences rather than perfect optimisation \cite{hofbauer2003evolutionary,hu2020optimal}, \cite{salahshour2025perceptual}.

The game is non-zero-sum, as the gain of one side is not necessarily equal to the loss of the other—both players may simultaneously gain or lose depending on their strategies \cite{park2025non}. The dynamics capture realistic cyber engagement, where defence improvements may reduce attack opportunities without eliminating attacker payoffs entirely. The payoff a player $X$ obtains in a pairwise interaction with the player $Y$ is defined in Table \ref{table:1}.  
\begin{table}[H]
   \centering
   \caption{Payoffs of the attacker and defender}
    \begin{tabular}{|c|c|c|c|}
      \hline
      \multicolumn{2}{|c|}{\textbf{Strategies}} & \multicolumn{2}{|c|}{\textbf{Payoffs}} \\
      \hline
     \textbf{D} & \textbf{A} & \textbf{Defender} & \textbf{Attacker}\\
     \hline
     $ND$ & $NA$ & 0 & 0 \\
     \hline
     $ND$ & $A$ & $-w$ & $-c_a+b_a$ \\
     \hline
     $D$ & $NA$ & $-c_d+b_d$ & 0 \\
     \hline
     $D$ & $A$ & $-c_d+p_db_d-w(1-p_d)$ & $-c_a+b_a(1-p_d)$ \\
     \hline
  \end{tabular}
    \label{table:1}
\end{table}
Payoffs represent the expected utility outcomes of cyber confrontation, incorporating factors such as defence success probability, cost and benefits of defence and attack, and damage resulting from failed defence. Parameter values and their meanings are given in Table \ref{table:2} and players' average payoffs are available in the Appendix in equations A1 to A4. 
 
\begin{table}[H]
    \centering
    \renewcommand{\arraystretch}{1.5}
        \caption{Summary of parameters in the model}
    \label{table:2}
    \begin{tabular}{|c|m{8cm}|c|}
    \hline
    \textbf{Variables} & \textbf{Meaning of variables} & \textbf{Constraints}\\
    \hline
    $w$  & Assets value, i.e. loss to the defender for an attack & $0 < w \leq1$ \\
     \hline
     $c_a$  & Cost to the attacker for attack attempt & $ 0< c_a < w $ \\
    \hline
    $c_d$  & Cost to the defender for implementing defence & $ 0 < c_d <w $ \\
    \hline
     $b_a$  & Attacker's benefit for a successful attack & $c_a < b_a$ \\
     \hline
    $b_d$  & Defender’s benefit for not being breached & $c_d < b_d \leq w$ \\
     \hline
     $p_d$  & Probability of successful defence & $0 < p_d\leq1$ \\
     \hline
    \end{tabular}
\end{table}

In the following subsection, we analyse the fixation and transition probabilities and stationary distribution of strategies and derive risk-dominant conditions for attack and defence to understand how the strategies evolve in a well-mixed finite population. We first analyse a baseline model in which defenders choose between defence $(D)$ and no-defence $(ND)$ without any external intervention. We then extend this framework in two steps: first by introducing committed defenders who always adopt defence, and subsequently by incorporating subsidy mechanisms that reduce the cost of defence. This progression allows us to isolate the impact of commitment and incentives on cybersecurity outcomes.

\subsection{Evolutionary dynamics in cyber security}

We model the dynamics using the frequency-dependent Moran process \cite{fudenberg2006,imhof2005evolutionary}, where strategy evolution follows a stochastic learning process rather than perfect rational optimisation. In this case, individuals adapt their behavioural strategy by imitating strategies that achieve higher payoffs, and therefore more successful individuals tend to be imitated more often than the others \cite{han2016emergence,traulsen2006stochastic}. 
To analyse the coupled attacker–defender dynamics, we approximate the evolutionary process as a Markov chain transitioning between four homogeneous states: $(A,D)$, $(NA,D)$, $(A,ND)$, and $(NA,ND)$. As state transition is driven by the imitation process, the evolutionary dynamics unfold in one population at a time. 
Therefore, we only analyse transitions between states that differ in the strategy of a single population. Transitions across both populations, such as from $(A,D)$ to $(NA,ND)$, occur with zero probability. 
To illustrate, consider the transition from $(A,D)$ to $(NA,D)$. Suppose that at a given time, the attacker population contains $k$ agents using strategy $A$ $(0 \leq k \leq N)$ and $(N-k)$ agents using strategy $NA$. Then the probability of $A$ switching to $NA$ when the defence population is in all-defence $(D)$ state follows the Fermi function \cite{traulsen2006stochastic}:

\begin{equation}\label{eq:1}
    p_{A,NA}^D = (1 + e^{-\beta(f_{NA}^D - f_A^D)})^{-1}
\end{equation}

Here, $\beta$ is the intensity of selection: $\beta = 0$ implies random imitation, while $\beta \rightarrow \infty$ leads to deterministic imitation of the better strategy. Using this update rule, the probability of changing the number of $i$ agents using strategy $A$ by $\pm 1$ in each time step can be written as: \cite{han2016emergence}:

\begin{equation}\label{eq:2}
    T^{\pm}(i) = \frac{N-i}{N}\frac{i}{N}\left[1+e^{{{\mp}}{\beta}[f_{A(i)}^D - f_{NA(i)}^D]}\right]^{-1}
\end{equation}

The fixation probability of a single mutant using a strategy $A$ introduced into a population of $(N - 1)$ agents using strategy $NA$ is given by:

\begin{equation}\label{eq:3}
    \rho_{NA,A}^D = \left(1+\sum_{k=1}^{N-1}\prod_{j=1}^{k} \frac{T^-(j)}{T^+(j)}\right)^{-1}
\end{equation}

We consider the rare mutation limit, where mutations occur infrequently so that each mutant either fixates or goes extinct before the next mutation appears \cite{veller2016finite,fudenberg2006imitation,nowak2004emergence}. 
This allows the dynamics to be approximated as a Markov chain over monomorphic states.
At each evolutionary step, only one population (either attackers or defenders) is assumed to update, while the other remains fixed. Therefore, transitions occur sequentially between populations, and each transition is weighted by a factor of $\frac{1}{2}$ to reflect the equal probability of updating either population.
The fixation probabilities computed above define the transition rates between the four pure states, forming the entries of a Markov transition matrix $M$ available in the Appendix (A5), where the states correspond to $x = \{(A,D),(NA,D),(A,ND),(NA,ND)\}$.

The long-term behaviour of the system is described by the stationary distribution $\pi$, defined as the normalised left eigenvector of $M$ associated with eigenvalue 1 \cite{imhof2005evolutionary}:
\[
\pi M = \pi, \quad \sum_{i=1}^{s} \pi_i = 1.
\]
Each $\pi_i$ represents the long-run probability that the population resides in state $i$. This stationary distribution allows us to identify which cyber-defence outcomes are most likely to emerge and persist under evolutionary dynamics.

\begin{figure}[H]
\centering
\setlength{\tabcolsep}{0pt}

\begin{subfigure}[t]{0.35\textwidth}
\centering
\includegraphics[width=\linewidth]{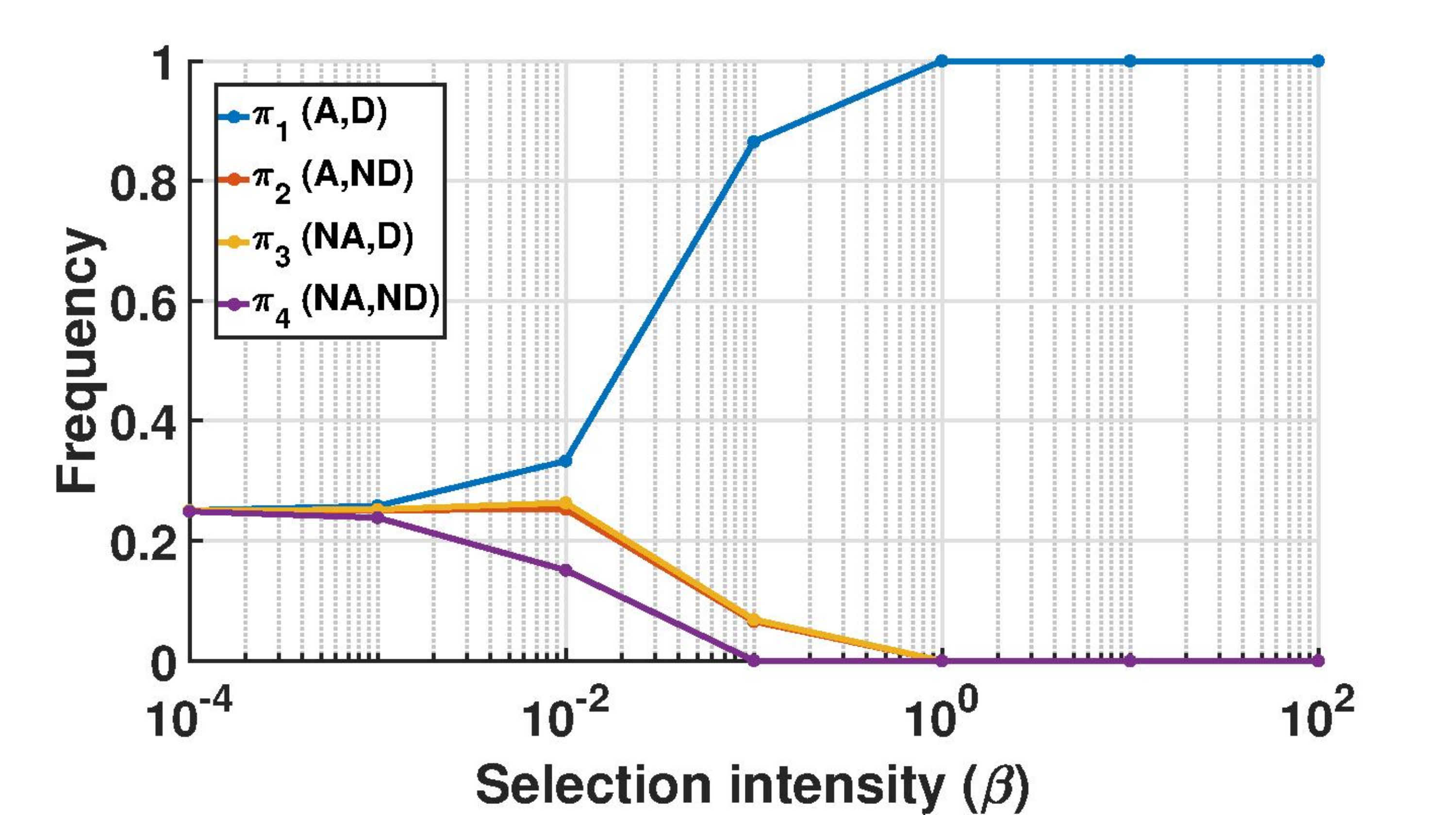}
\caption{}
\end{subfigure}\hspace{-6mm}
\begin{subfigure}[t]{0.35\textwidth}
\centering
\includegraphics[width=\linewidth]{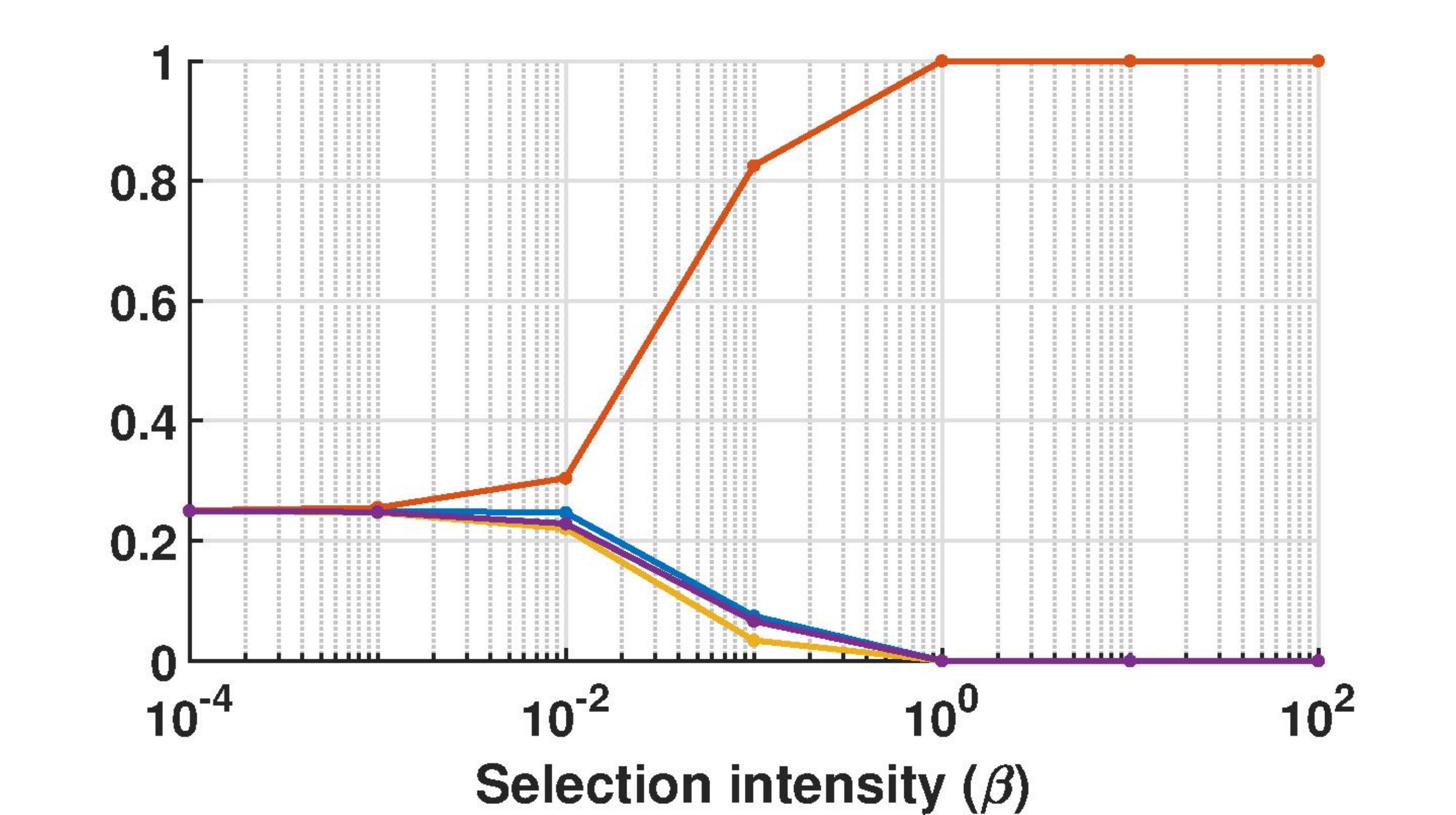}
\caption{}
\end{subfigure}\hspace{-6mm}
\begin{subfigure}[t]{0.35\textwidth}
\centering
\includegraphics[width=\linewidth]{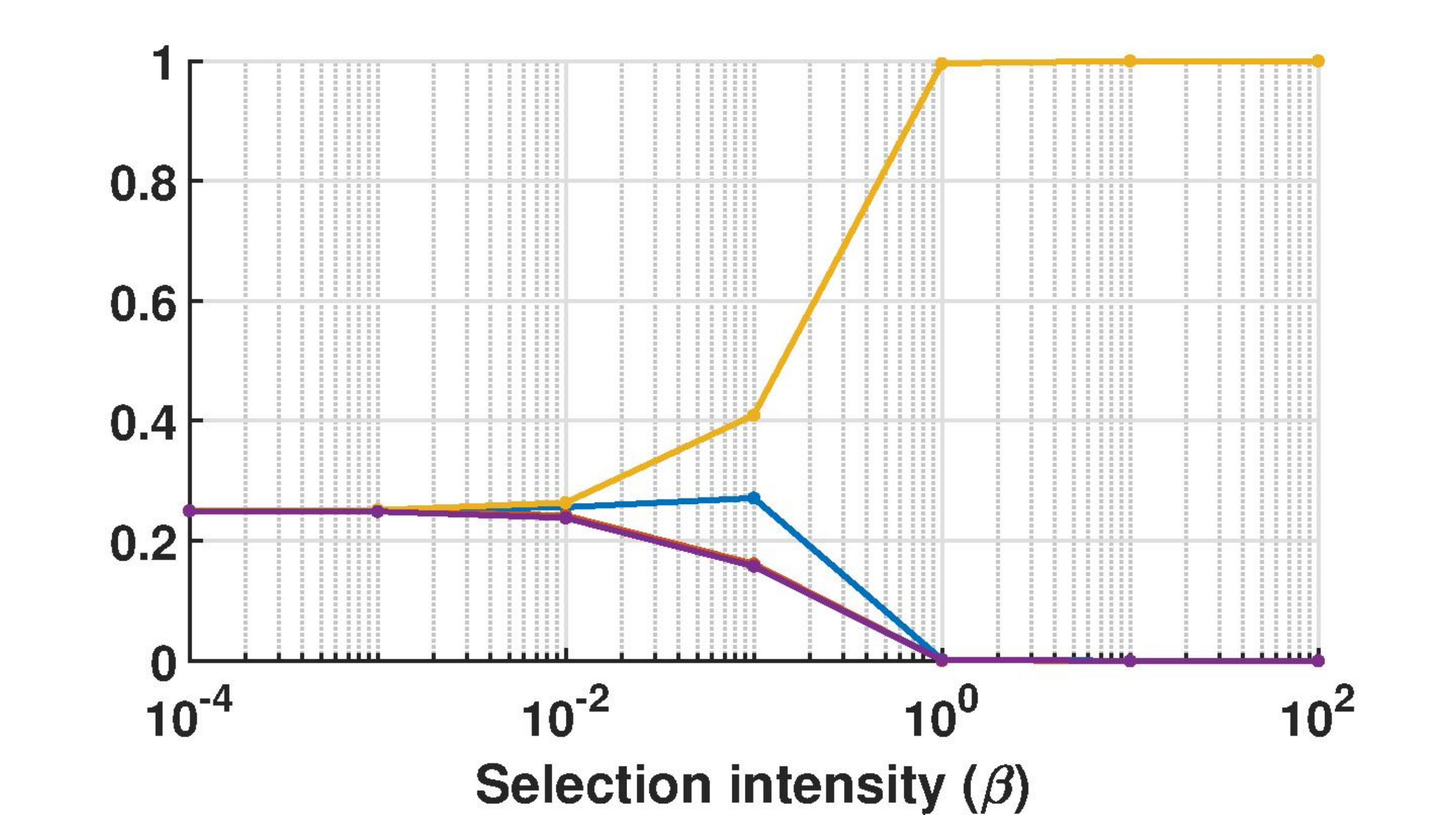}
\caption{}
\end{subfigure}

\begin{subfigure}[t]{0.30\textwidth}
  \centering
  \includegraphics[width=\linewidth]{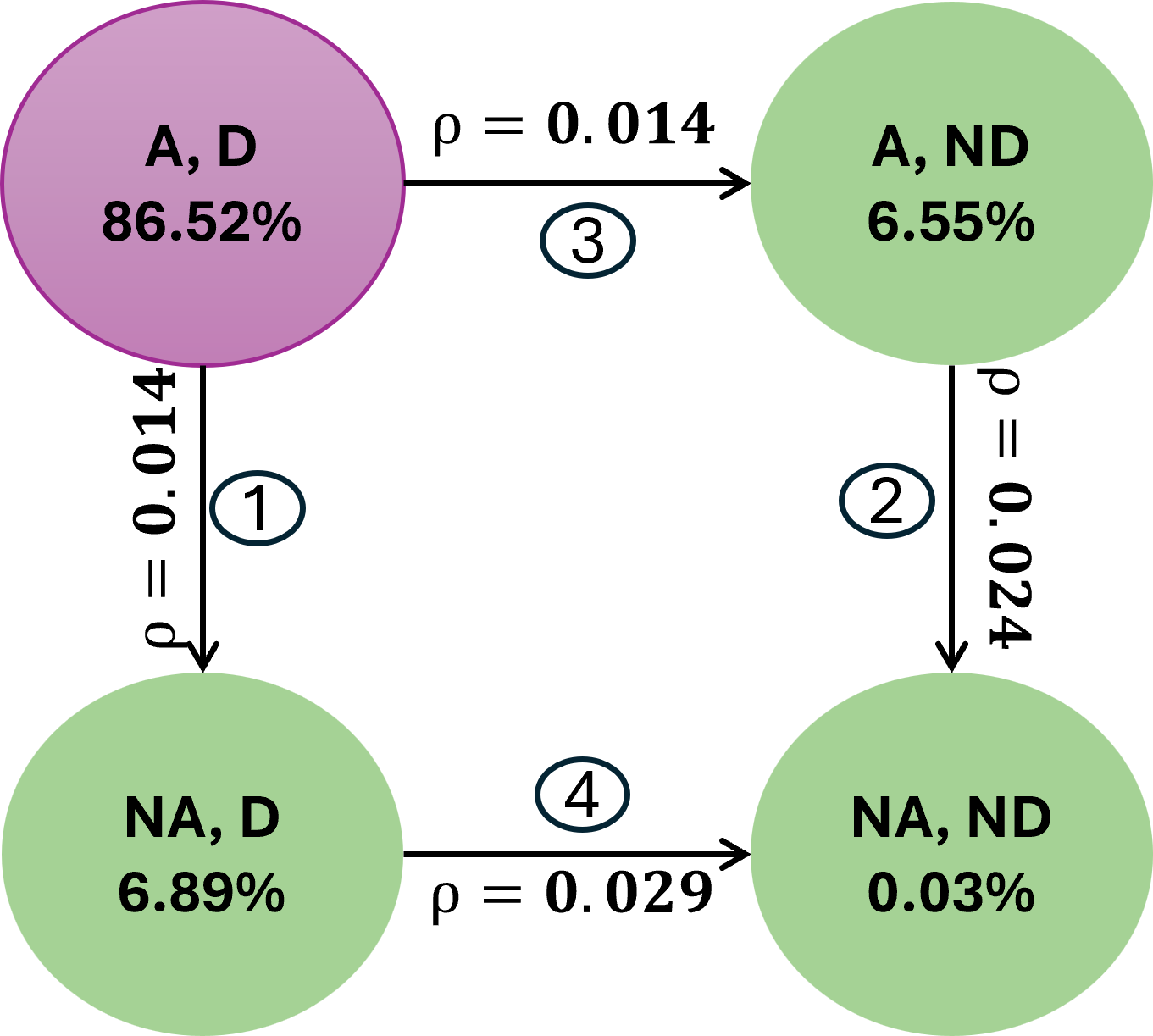}
  \caption{}
\end{subfigure}\quad
\begin{subfigure}[t]{0.30\textwidth}
  \centering
  \includegraphics[width=\linewidth]{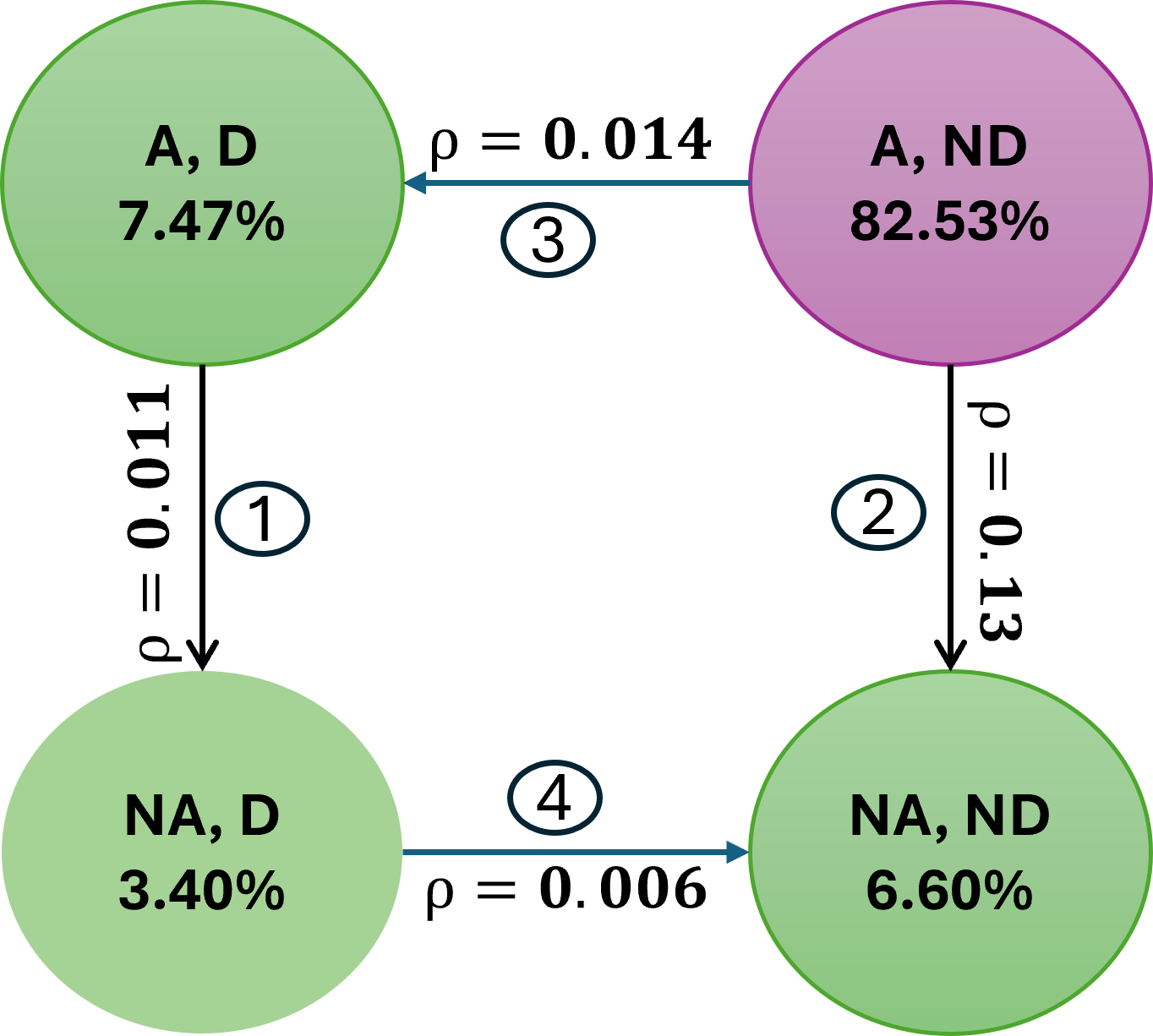}
  \caption{}
\end{subfigure}\quad
\begin{subfigure}[t]{0.30\textwidth}
  \centering
  \includegraphics[width=\linewidth]{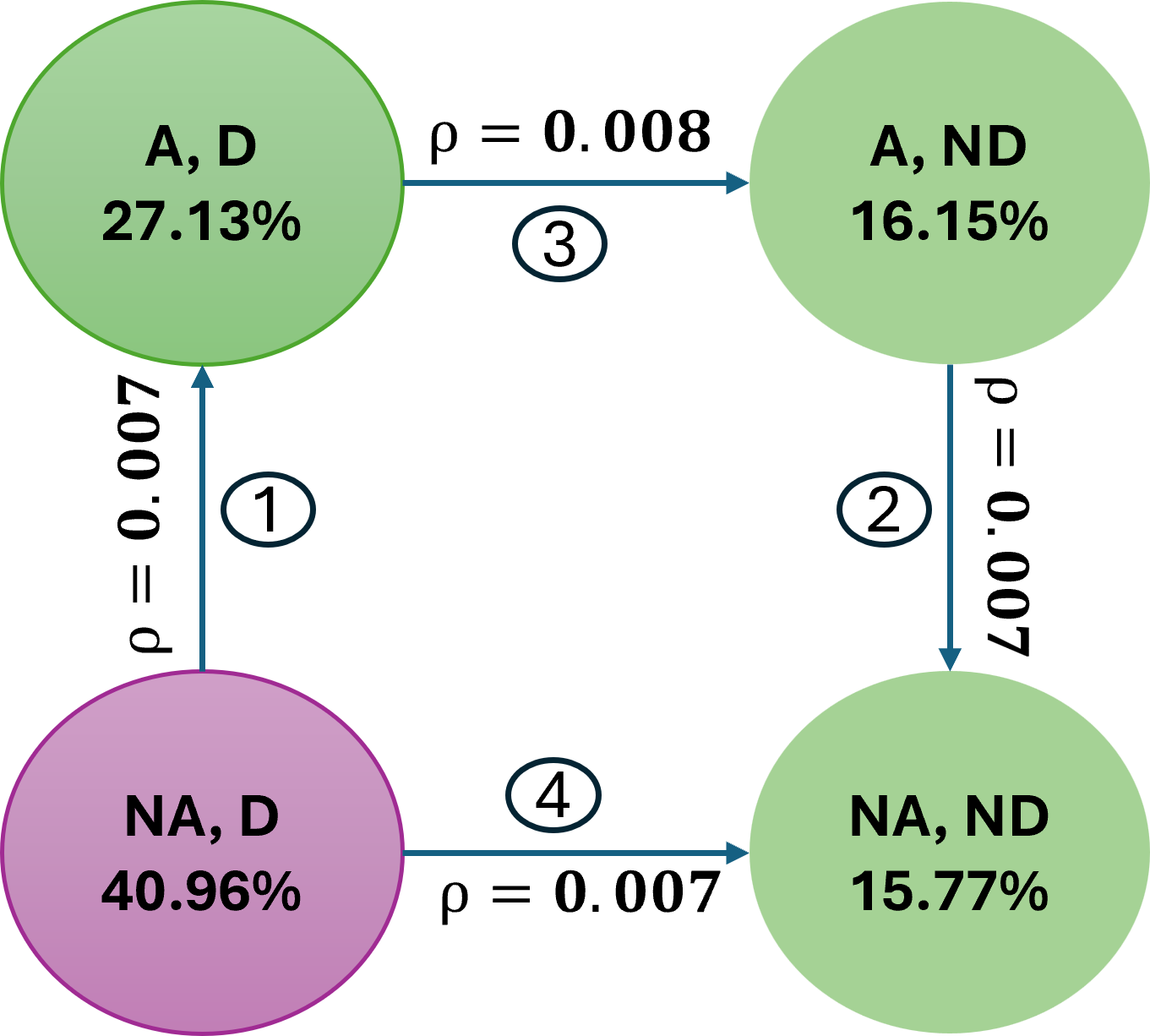}
  \caption{}
\end{subfigure}
\caption{Finite population analysis of cybersecurity. Each row corresponds to one representative parameter setting. Top row: stationary probabilities of the strategies $(A,D)$, $(A,ND)$, $(NA,D)$ and $(NA,ND)$ as functions of the selection intensity $\beta$. Bottom row: the corresponding embedded four-state Markov chains for $\beta=0.1$. 
Parameter values for $(a,d)$ are $b_a=0.90$, $b_d=0.79$, $c_a=0.41$, $c_d=0.20$, $w=0.98$, $p_d=0.26$ where attack and defence $(A,D)$ becomes dominant under strong selection. 
For $(b,e)$ parameter values are $b_a=0.52$, $b_d=0.37$, $c_a=0.29$, $c_d=0.34$, $w=0.43$, $p_d=0.09$ where attack and not defence $(A,ND)$ dominates. 
For $(c,f)$ parameter values are $b_a=0.24$, $b_d=0.47$, $c_a=0.18$, $c_d=0.41$, $w=0.47$, $p_d=0.54$ where not attack and defence $(NA,ND)$ dominates. 
Node colour (purple) indicates the dominant strategy pair. The numbers in the oval shape show the risk-dominant strategy from Table \ref{tab:3}. Only transitions where the probability is larger than the reverse direction are shown. Node labels give the stationary distribution of each state, and edge labels show fixation probabilities $\rho$. The dominant outcome in each case is determined by the relative payoff advantage of attack versus defence, illustrating how cost and effectiveness parameters shape evolutionary stability.}
\label{fig:1}
\end{figure}

Figure \ref{fig:1} illustrates the baseline finite-population dynamics of the four strategy pairs. Across the three parameter settings, the stationary distributions are shown in subplots $(a)$–$(c)$ to show how payoff differences determine the long-run outcome. The corresponding four-state Markov chains in $(d)$–$(f)$ illustrate the fixation pathways between strategy pairs, showing how payoff differences drive the system toward specific dominant states. 
When the attacker payoff exceeds the cost of attack (i.e., $b_a(1-p_d) > c_a$) and defence is costly, attack without defence $(A,ND)$ becomes dominant, as shown in $(b)$ and $(e)$. In contrast, when defence is sufficiently effective and its cost is justified by the benefit (i.e., $p_d b_d > c_d$), the system converges to defence-dominant states such as $(NA,D)$, as in $(c)$ and $(f)$. When neither condition strongly dominates, the system exhibits mixed behaviour and $(A,D)$ dominates, as seen in $(a)$ and $(d)$.
These results are consistent with the analytical conditions derived from the payoff structure, where dominance is determined by the relative cost–benefit trade-offs of attack and defence strategies.

\begin{table}[H]
\centering
\small
\setlength{\tabcolsep}{6pt}
\renewcommand{\arraystretch}{1.2}
\caption{Risk-dominant conditions for the attack--defence model.}
\label{tab:3}
\begin{tabular}{|c| l l |c| l l|}
\hline
\textbf{\#} & \textbf{Attacker transition} & \textbf{Condition} & \textbf{\#} & \textbf{Defender transition} & \textbf{Condition} \\
\hline
1 & $(A,D)\rightarrow(NA,D)$   & $c_a > b_a(1-p_d)$ 
  & 3 & $(A,D)\rightarrow(A,ND)$   & $c_d > p_d(b_d + w)$ \\
\hline
2 & $(A,ND)\rightarrow(NA,ND)$ & $c_a > b_a$ 
  & 4 & $(NA,D)\rightarrow(NA,ND)$ & $c_d > b_d$ \\
\hline
\end{tabular}
\end{table}

In Markov diagrams, arrow directions show the risk-dominant conditions given in Table \ref{tab:3}, while the displayed edge weights give the corresponding transition probabilities. Risk-dominant conditions show that attack/defence is favoured in the presence of defenders/attackers if the benefits are higher than the cost to the players. Therefore, the driving factor of $A$ and $D$ strategies is the cost incured by the players. Together, these results demonstrate how finite populations respond to payoff asymmetries and provide the foundation for our broader parameter-sweep analysis, where we systematically explore that besides cost, how varying benefits, and defence effectiveness reshape the evolutionary landscape.

\begin{figure}[H]
\centering
\includegraphics[width=1.03\linewidth]{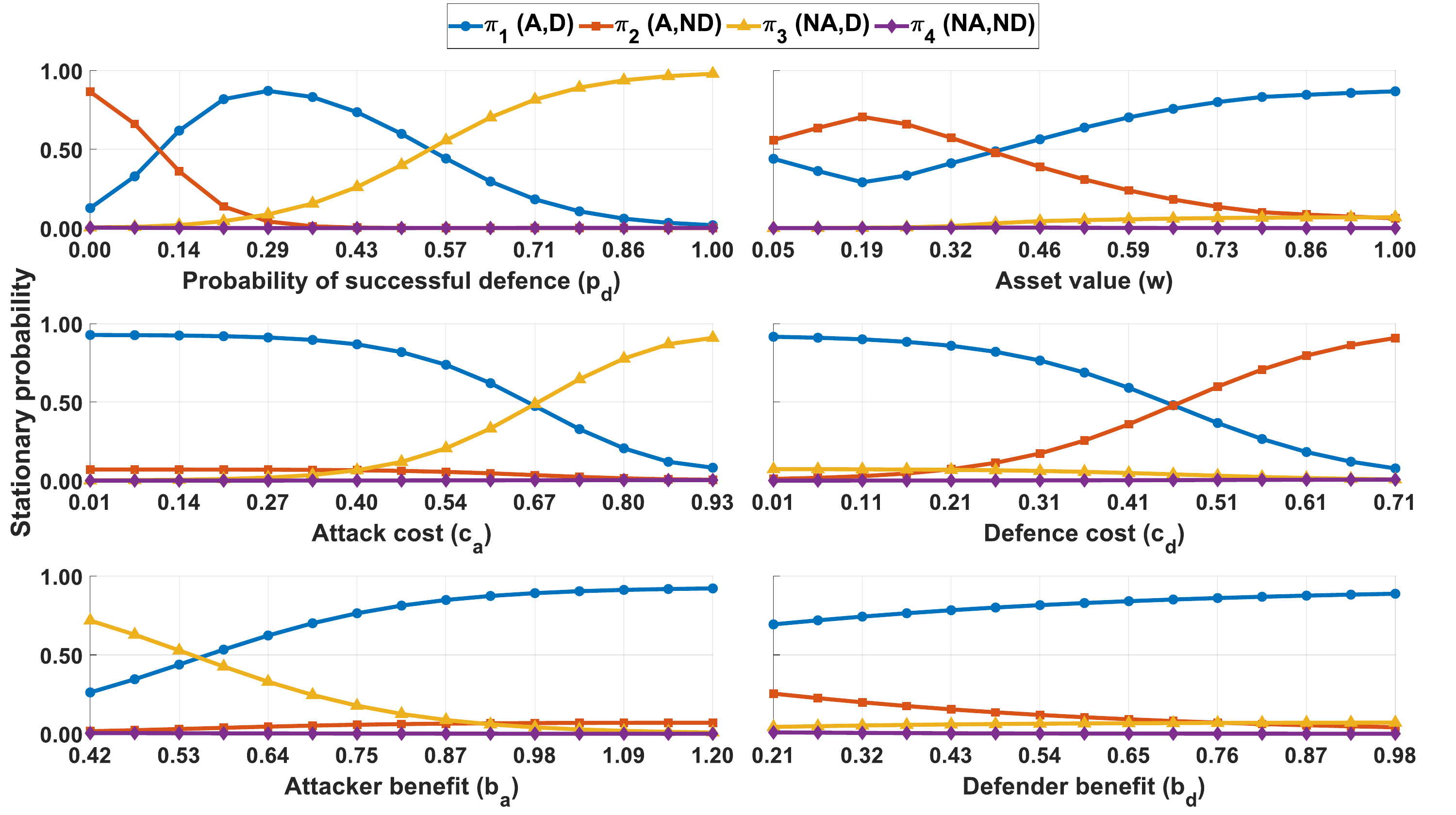}
\caption{Frequency of strategies with respect to the parameters such as defence probability $(p_d)$, assest value $(w)$, cost of attack $(c_a)$ and defence $(c_d)$, and benefit of attack and defence $(b_a)$, $(b_d)$, respectively. Parameter values are $N=100$, $b_a=0.90$, $b_d=0.79$, $c_a=0.41$, $c_d=0.20$, $w=0.98$, $p_d=0.26$ and $\beta=0.1$. The results highlight that higher $(p_d)$ and $(b_d)$ increase the likelihood of defence strategies, while attack is avoidable by increasing $(c_a)$, and $(p_d)$.}
\label{fig:2}
\end{figure}

\paragraph{Parameter-Driven Behaviour Insights.}

To understand how cyber–attack and defence strategies evolve under different security conditions, we conduct parameter sweeps and analyse their effect on the stationary distribution $\pi$. The results in Figure~\ref{fig:2} illustrate how system dynamics shift between attack-dominated and defence-dominated equilibria depending on payoff parameters. First, when $p_d < 0.3$, defence is unreliable and attacking strategies $(A,D)$ and $(A,ND)$ dominate. As $p_d$ increases ($p_d \approx 0.3$–$0.5$), attackers lose their payoff advantage and $(NA,D)$ rises. For $p_d > 0.55$, the system rapidly converges to $(NA,D)$, effectively driving attacks to extinction. A similar transition is observed with the loss parameter $w$. When $w < 0.3$, the penalty of failed defence is small, allowing attackers to exploit high-value defence targets, so $(A,D)$ dominates. As $w$ increases ($w \approx 0.3$–$0.5$), a transition region emerges, and for $w > 0.5$, defenders strongly prefer high defence, leading to stable $(NA,D)$. The cost of attacking also plays a critical role. When $c_a < 0.2$, attacking remains profitable and $(A,D)$ dominates. As $c_a$ increases ($c_a \approx 0.3$–$0.5$), attacks become less attractive and $(NA,D)$ gains probability. For $c_a > 0.55$, attacks collapse entirely. Similarly, defence cost affects stability. When defence is cheap ($c_d < 0.2$), defenders maintain high-defence effort and $(NA,D)$ prevails. Moderate cost ($c_d \approx 0.3$–$0.45$) introduces instability, while for $c_d > 0.5$, defenders shift to low defence and attack strategies re-emerge. The benefit parameters further shape the dynamics. For $b_a < 0.4$, attacks provide little reward and $(NA,D)$ dominates, whereas increasing $b_a$ ($b_a \approx 0.4$–$0.7$) introduces coexistence. For $b_a > 0.8$, high rewards drive attack resurgence and $(A,D)$ dominates. In contrast, strong defence incentives ($b_d > 0.6$) make defence valuable, leading to $(NA,D)$ dominance, whereas when $b_d < 0.4$, defending yields weak payoff and the attack-tolerant state $(A,ND)$ emerges. 

Overall, these results demonstrate that the non-attacking, always-defence strategy $(NA,D)$ becomes the dominant stationary outcome whenever
\[
p_d > 0.5, \quad w > 0.5, \quad c_a > 0.4, \quad c_d < 0.3, \quad b_a < 0.6, \quad b_d > 0.6 .
\]

Under these conditions, attacks lose evolutionary fitness and the population self-organises into a stable defence equilibrium. Below these thresholds, the system shifts sharply toward attack-dominated regimes. These findings are further supported by robustness analysis across $10{,}000$ randomly generated games in Figure~\ref{fig:3}, confirming a broad parameter region where $(NA,D)$ dominates and successful attacks remain rare. We computed the stationary distribution for each sample, and the results were averaged across all instances. In other words, defence stabilises and attacks remain rare whenever (i) defence is strong, (ii) attacking well-defended systems is expensive, (iii) defence prevents substantial losses, and (iv) the benefit from a successful attack is not too high. Building on these insights that defence stability depends on the capability of defenders, we now introduce a differential access framework that strategically allocates advanced AI security tools across the defender population.
The baseline analysis yields the following key insights. 
\textbf{(i)} High defence cost discourages adoption of defence, driving the system toward weak defence and sustained attack activity. 
\textbf{(ii)} Defence becomes dominant only when it is sufficiently effective and economically viable, highlighting the importance of cost–benefit balance.}

\begin{figure}[H]
\centering
\includegraphics[width=\linewidth,height=0.55\textheight,keepaspectratio]{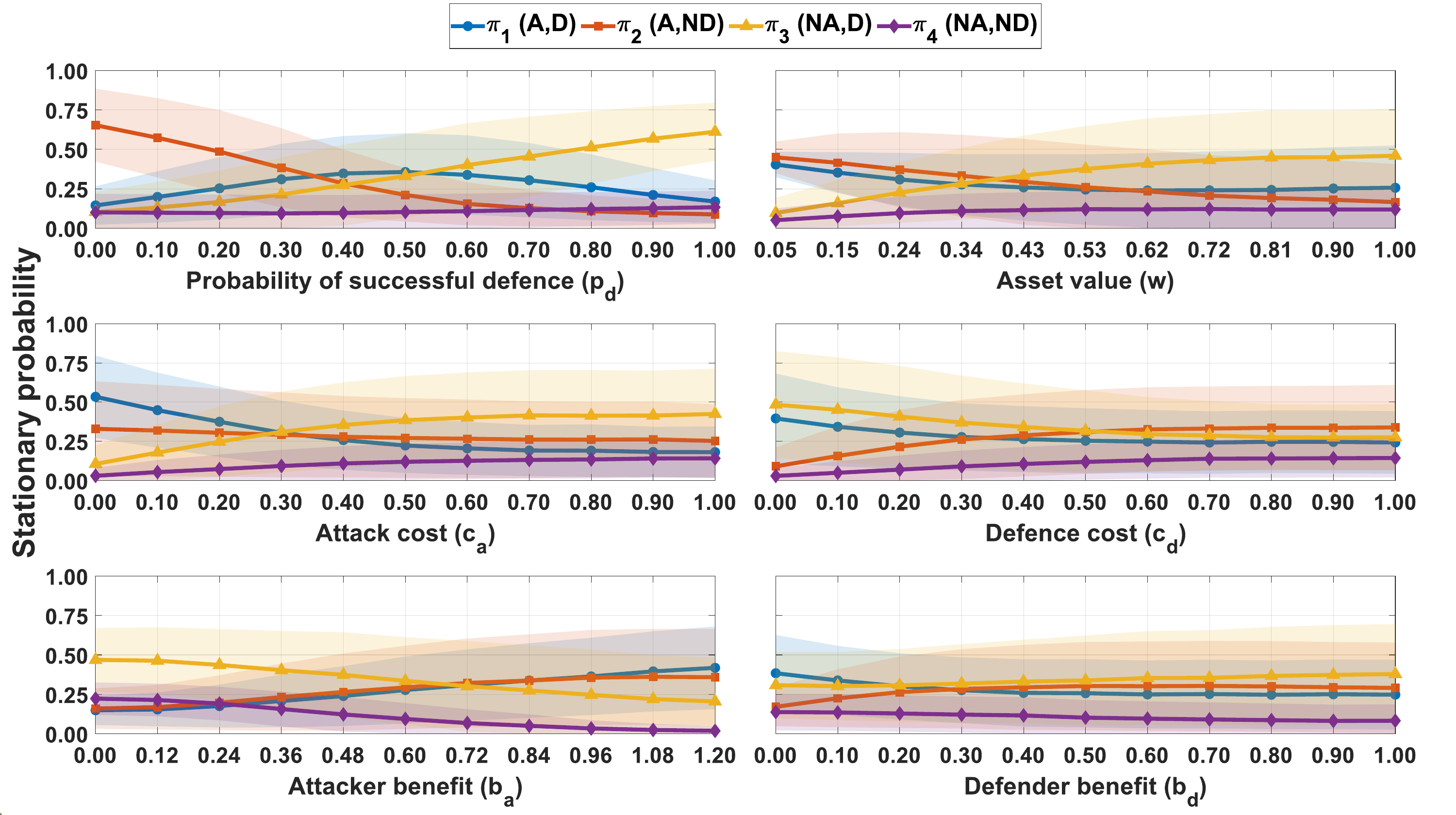}
\caption{Robustness of stationary strategy frequencies under parameter variation  ($\beta = 1$). Each subplot shows the mean stationary probability of the four strategy states as a single model parameter is varied while all other parameters are fixed at baseline values from Figure \ref{fig:2}. The solid lines indicate the mean probability, and the shaded regions represent $\pm$ one standard deviation, capturing variability from 10,000 randomly generated  games, for robustness check. Narrow shaded bands indicate robustness (stable outcomes despite parameter perturbations), while wider bands reflect sensitivity to the chosen parameter.}
\label{fig:3}
\end{figure}

\section{Strategic AI Access Allocation}
\label{4}

The parameter-driven analysis in the previous section shows that stable defence states emerge only when defence is sufficiently effective, affordable, and rewarding. However, in real cyber ecosystems these favourable conditions rarely hold. Many organisations, particularly SMEs, cannot afford advanced security infrastructure or specialised staff, despite facing a high share of attacks \cite{wilson2025one,arroyabe2024revealing}. As a result, defenders often adopt minimal low-cost protection, facing persistent attack success and preventing convergence to a secure equilibrium. 

\begin{figure}[ht]
    \centering
    \includegraphics[width=0.95\linewidth]{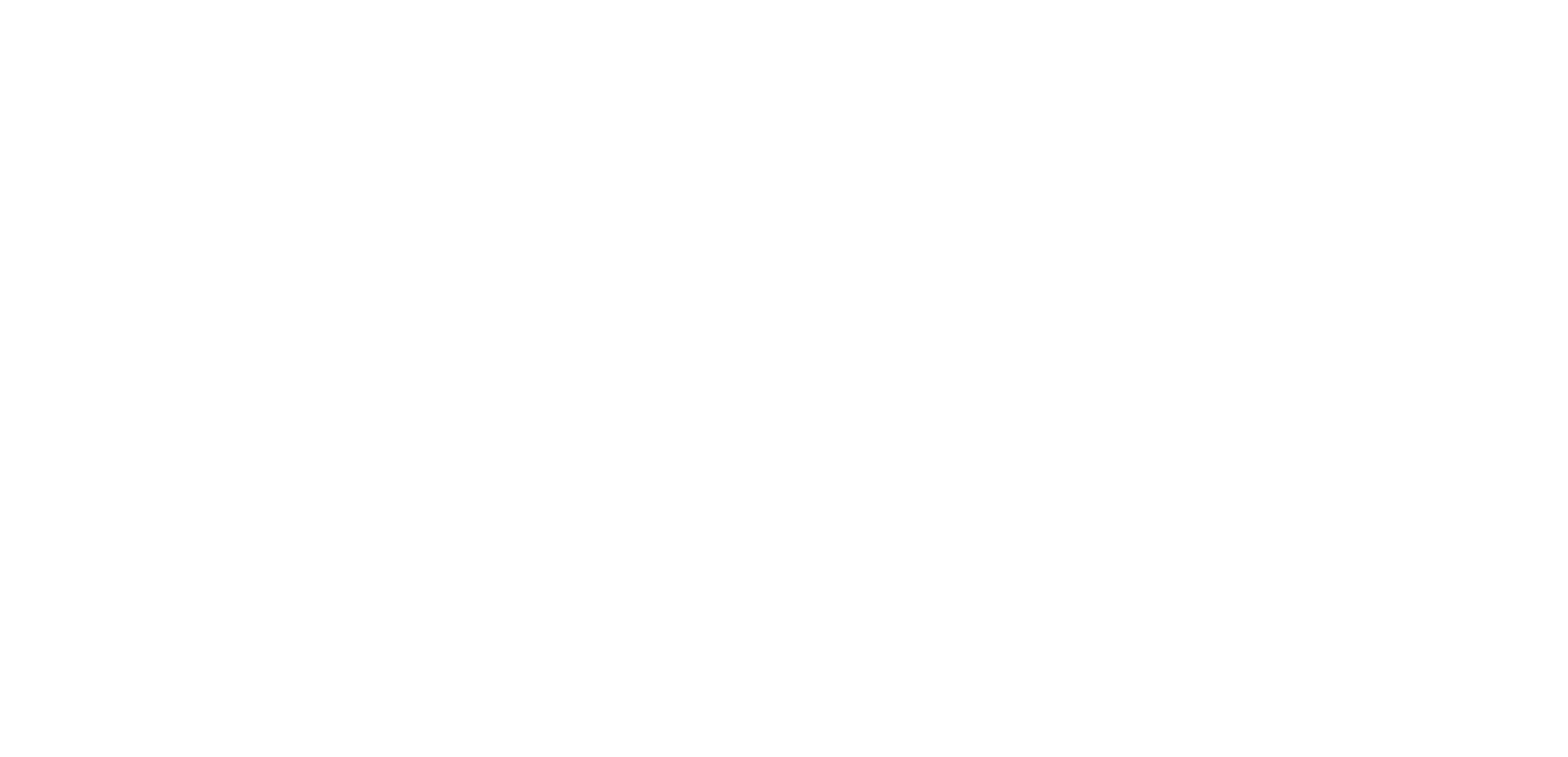}
    \caption{\textbf{Graphical illustration of the proposed cyber defence model with differential access, committed defenders, and subsidy.} The attacker population consists of agents choosing between attacking (A) and non-attacking (NA), while the defender population consists of agents choosing high defence (H-D) or low defence (L-D). Dashed lines represent social imitation, where agents copy strategies with higher payoffs, and arrows indicate the evolutionary transitions over time. Initially, high successful attacks lead to weak defence (ND/D). With differential access, defenders can adopt high (H-D) or low (L-D) defence, but high defence incurs a significant cost. The introduction of committed defenders ($z$), who always choose high defence, increases the adoption of H-D strategies; however, the high cost limits improvements in social welfare. Finally, under subsidy ($C_H = 0$), the cost of high defence is removed, leading to greater adoption of H-D, fewer successful attacks, and higher social welfare. Overall, the attacker and defender populations co-evolve and continuously influence each other’s strategy dynamics.}
    \label{fig:4}
\end{figure}

The baseline model in Section \ref{3} assumes uniform access to defence and therefore cannot capture the impact of unequal resource availability across defenders. In practice, defenders vary significantly in their ability to deploy advanced AI-based security tools. This gap motivates the need for mechanisms that actively shape defender behaviour rather than relying solely on market-driven adoption. A promising direction, recently highlighted in AI governance and safety policy debates, is the principle of \textit{differential access} to AI security tools \cite{ee2025asymmetry,hammerschmidt2025bridging}. Rather than providing identical capabilities to all players, access to high-assurance AI defence tools may be restricted to trusted or regulated defenders who can operate them responsibly. In practice, advanced AI defence platforms (e.g., policy-compliant LLMs, automated threat–response systems, and protected inference environments) are increasingly deployed under controlled access conditions \cite{OpenAI2025ChatGPTFederal}. This reflects growing recognition that not every entity should hold unrestricted high-capability tools, and that selective empowerment may improve collective resilience \cite{seger2023open,ilcic2025artificial}.

To capture this unequal access to resources, we generalise the baseline model by introducing heterogeneous levels of defensive capability. In particular, we divide defenders into two strategic groups: \textbf{High-access} $(H)$ defenders, equipped with full-spectrum secure AI capabilities (e.g., automated patching, predictive threat modelling), correspond to the defence (D) strategy, and \textbf{Low-access} $(L)$ defenders, who rely on basic public tools with limited performance, corresponds to the no-defence (ND) strategy. The key difference is that we now explicitly distinguish between different levels of effectiveness and cost, allowing us to model realistic inequalities in cyber readiness. While prior work proposes memory-based and recovery mechanisms to mitigate attacks, such approaches do not address the economic barriers that limit adoption of strong defence strategies \cite{hu2025security}. Our model creates an asymmetric resource-allocation game in which defenders select their defence level based on expected payoff, cost, and risk exposure to capture the strategic dilemma of whether to invest in costly robust defence or adopt cheaper but weaker protection.

The mathematical model below formalises these interactions. The payoffs for attacker and non-attacker in the presence of $H$ defence are $f_A^H$ and $f_{NA}^H$, respectively. Similarly, the payoffs when interacting with $L$ defence are $f_A^L$ and $f_{NA}^L$.

\noindent\textbf{Attacker payoffs:}
\begin{equation}\label{eq:4}
\begin{split}
    f_A^H   &= -c_a + b_a\,(1-p_{dH}), \quad
    f_A^L   = -c_a + b_a\,(1-p_{dL}),\\
    f_{NA}^H  &= 0, \quad 
    f_{NA}^L  = 0.
\end{split}
\end{equation}

\noindent Similarly, the payoffs for $H$ defender and $L$ defender in the presence of $A$ and $NA$, are respectively:

\noindent\textbf{Defender payoffs:}
\begin{equation}\label{eq:5}
\begin{split}
    f_H^A   &= p_{dH}B_H - C_H - (1-p_{dH})W_H, \quad
    f_L^A   = p_{dL}B_L - C_L - (1-p_{dL})W_L,\\
    f_H^{NA}  &= B_H - C_H, \quad
    f_L^{NA}  = B_L - C_L.
\end{split}
\end{equation}

The parameters are consistent with those introduced in the baseline model. In particular, $c_a$ and $b_a$ denote the attack cost and benefit, respectively. The key extension is that defence effectiveness now depends on access level, captured by $p_{dH}$ and $p_{dL}$ for high- and low-access defenders. Similarly, $B_H$ and $B_L$ denote the benefits of maintaining high- and low-level defence, $C_H$ and $C_L$ represent the corresponding defence costs, and $W_H$ and $W_L$ capture the losses from successful attacks. 
We assume that high defence incurs a higher cost than low defence ($C_H > C_L$), reflecting the expense of advanced AI-based security tools. Similarly, we assume $p_{dH} > p_{dL}$ as increasing AI defence capability can increase the success probability. $B_H > B_L$ is assumed because better defence tools can be more beneficial, as they can deter attacks and reduce defender loss. These parameters collectively model realistic trade-offs between security effectiveness and economic cost.

\begin{table}[H]
\centering
\renewcommand{\arraystretch}{1.5}
\caption{Payoff structure for attacker and defender strategies.}
\label{table:3}
\begin{tabular}{|c|c|c|}
\hline
 & \textbf{High Defence $(H)$} & \textbf{Low Defence $(L)$} \\ \hline
\textbf{Attack $(A)$} & $(f_A^H, f_H^A)$ & $(f_A^L, f_L^A)$ \\ \hline
\textbf{Not Attack $(NA)$} & $(f_{NA}^H, f_H^{NA})$ & $(f_{NA}^L, f_L^{NA})$\\
\hline
\end{tabular}
\end{table}

To encourage adoption of $H$ defence strategies, we incorporate a subset of \textit{committed $H$ defenders} (zealots) who always adopt $H$ defence and never change their strategy through social learning.  In the literature, empirical and theoretical work shows that even a small number of committed individuals can shift evolutionary dynamics and drive prosocial  behaviours such as cooperation, fairness and safe AI development, to fixation \cite{masuda2012evolution,zimmaro2024emergence,cimpeanu2022artificial,santos2019evolution}. In cyber defence, these zealots represent government bodies, defence vendors, and regulated service providers who act as stable anchors for security best practice.

To model strategy adoption in a population size $N$, we consider $z$ committed $H$ defenders, and $N-z$ ordinary defenders who learn and update their strategy via social learning. The defender and attacker average payoffs become:

\begin{equation}\label{eq:6}
\begin{split}
    \Pi_H &= \frac{(m_A)f_H^A+(N-m_A)f_H^{NA}}{N}, \qquad
    \Pi_L = \frac{(m_A)f_L^A+(N-m_A)f_L^{NA}}{N},\\
    \Pi_A &= \frac{(m_H+z)f_A^H+(N-z-m_H)f_A^L}{N}, \qquad
    \Pi_{NA} = 0.
\end{split}
\end{equation}

\noindent Strategy updates follow a pairwise comparison (Fermi) rule \cite{traulsen2007pairwise}, as in equation \ref{eq:1}, and evolve via a frequency-dependent birth–death process.
Transition probabilities and fixation dynamics are given by:
\begin{equation}\label{eq:7}
    T_{n_H}^{+} = \frac{N-z-n_H}{N-z}\frac{n_H+z}{N}p_{L,H}^A, \qquad
    T_{n_H}^{-} = \frac{n_H}{N-z}\frac{N-z-n_H}{N}p_{H,L}^A,
\end{equation}
and for attackers:
\begin{equation}\label{eq:8}
    T_{n_A}^{+} = \frac{N-n_A}{N}\frac{n_A}{N}p_{NA,A}^D, \qquad
    T_{n_A}^{-} = \frac{n_A}{N}\frac{N-n_A}{N}p_{A,NA}^D.
\end{equation}
The fixation probability of a single mutant with a strategy $H$ in a defender population of (N-z-1) agents using $L$, when other population is all attack is given by:
\begin{equation}\label{eq:9}
    \rho_{L,H}^A = \left(1+\sum_{k=1}^{N-z-1}\prod_{j=1}^{k} \frac{T_{n_H}^-(j)}{T_{n_H}^+(j)}\right)^{-1}
\end{equation}
The fixation probability for the attacker population (N), when the other population is all defence:
\begin{equation}\label{eq:10}
    \rho_{NA,A}^D = \left(1+\sum_{k=1}^{N-1}\prod_{i=1}^{k} \frac{T_{n_A}^-(i)}{T_{n_A}^+(i)}\right)^{-1}
\end{equation}
The stationary distribution $\pi$ satisfies $\pi M=\pi$. To complement analytical results, we run stochastic simulations exploring how changes in cost and selection intensity alter adoption dynamics and social welfare.

\begin{table}[H]
\centering
\small
\setlength{\tabcolsep}{8pt}
\renewcommand{\arraystretch}{1.15}
\caption{Risk-dominant conditions for the differential-access model.}
\label{tab:6}
\begin{tabular}{|c| l| l|}
\hline
\textbf{\#} & \textbf{Risk-dominant arrow} & \textbf{Condition}\\
\hline
1 & $(A,H)\ \rightarrow\ (NA,H)$  & $c_{ah} > b_{ah}(1-p_{dH})$\\
\hline
2 & $(A,L)\ \rightarrow\ (NA,L)$  & $c_{al} > b_{al}(1-p_{dL})$\\
\hline
3 & $(A,H)\ \rightarrow\ (A,L)$  & $p_{dH}B_H - C_H - (1-p_{dH})W_H < p_{dL}B_L - C_L - (1-p_{dL})W_L$\\
\hline
4 & $(NA,H)\ \rightarrow\ (NA,L)$ & $B_H - C_H < B_L - C_L$\\
\hline
\end{tabular}
\end{table}

The risk-dominant conditions in Table~\ref{tab:6} identifies the conditions under which attack, no-attack, high defence, or low defence strategies become risk-dominant, based on the trade-off between attack benefits, defence effectiveness, and defence costs under differential access. Conditions (1)–(4) describe that attacks are discouraged when the attack cost exceeds the expected benefit adjusted by defence success probability. Conditions (5)–(8) capture defenders’ incentives to switch between $H$ and $L$ defence. These depend on the trade-off between defence effectiveness (reduced damage and higher protection probability) and defence cost. 

Overall, the table shows that high defence becomes risk-dominant only when its benefits outweigh its higher cost, while low defence persists when cost savings dominate. This highlights the central tension in the model: although $H$-defence is more effective, its adoption is limited by cost, which motivates the need for policy interventions such as subsidies.
  
The aim of the proposed approach, shown in Figure \ref{fig:4}, is to shift the cybersecurity balance in favour of defenders by finding the conditions to motivate them to adopt $H$ defence by using safe AI tools \cite{volk2024safer,emehin2024enhancing} and improve overall safety. A recent real-world example of differential AI access \cite{OpenAI2025ChatGPTFederal}, is the U.S. General Services Administration’s agreement with OpenAI to provide the federal workforce with ChatGPT Enterprise, for a nominal cost of $\$1$ per agency for one year. These tools were assessed on the grounds of security and performance and are free from ideological bias \cite{Bloomberg2025AIVendors}. It is a restricted, security-compliant version of the AI tool designed to protect sensitive information and meet strict compliance standards, to enhance defence effectiveness without compromising sensitive data. This selective provision of a safeguarded AI environment represents a controlled high-access scenario in our model, where defenders gain advanced defence capabilities to ensure cyber resilience.

\subsection{Population Dynamics with Differential Access}

\begin{figure}[ht]
\centering
\begin{subfigure}[t]{0.60\textwidth}
    \centering
    \includegraphics[width=\linewidth]{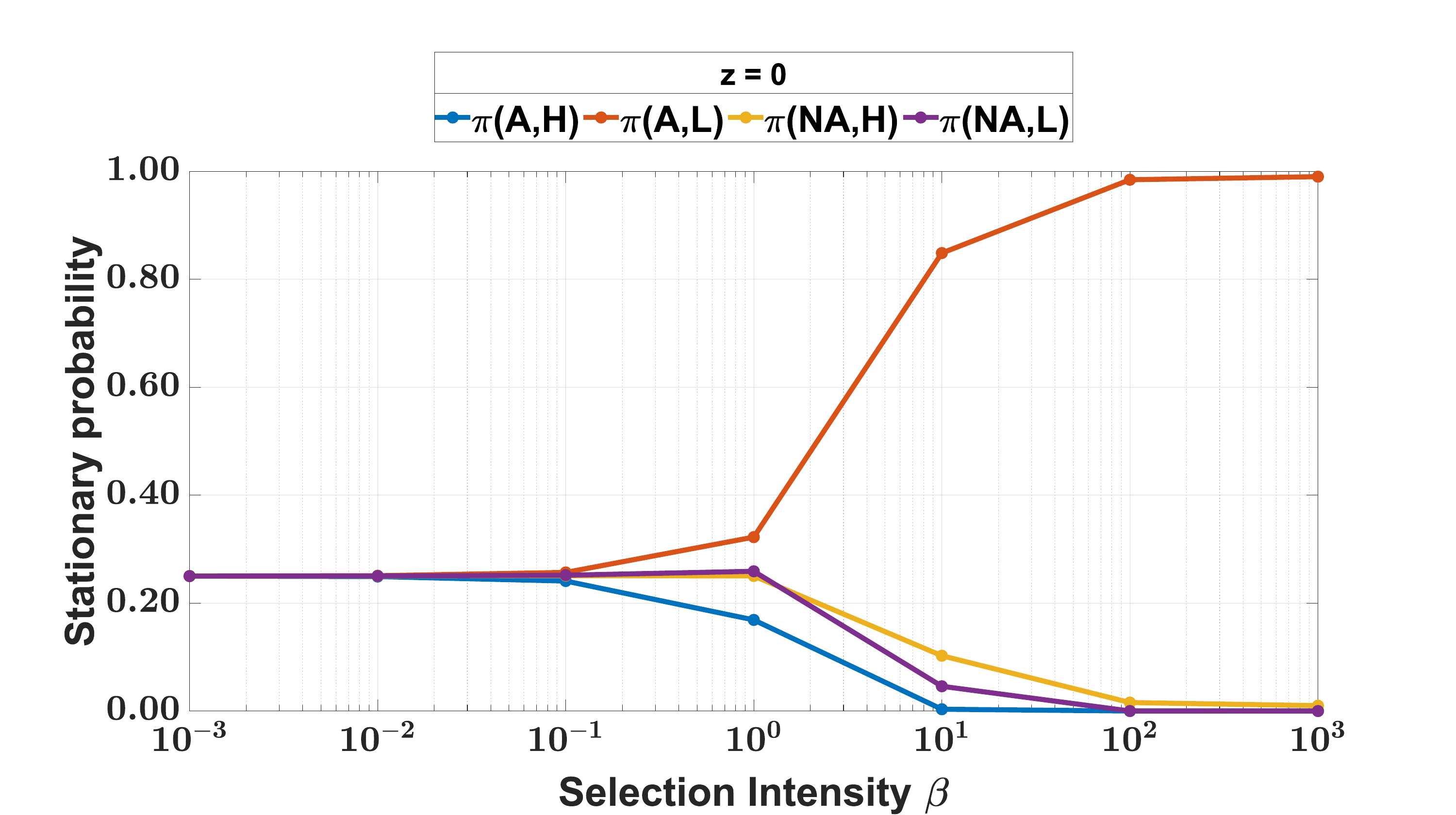}
    \caption{}
  \end{subfigure}
\begin{subfigure}[t]{0.38\textwidth}
    \centering
    \includegraphics[width=\linewidth]{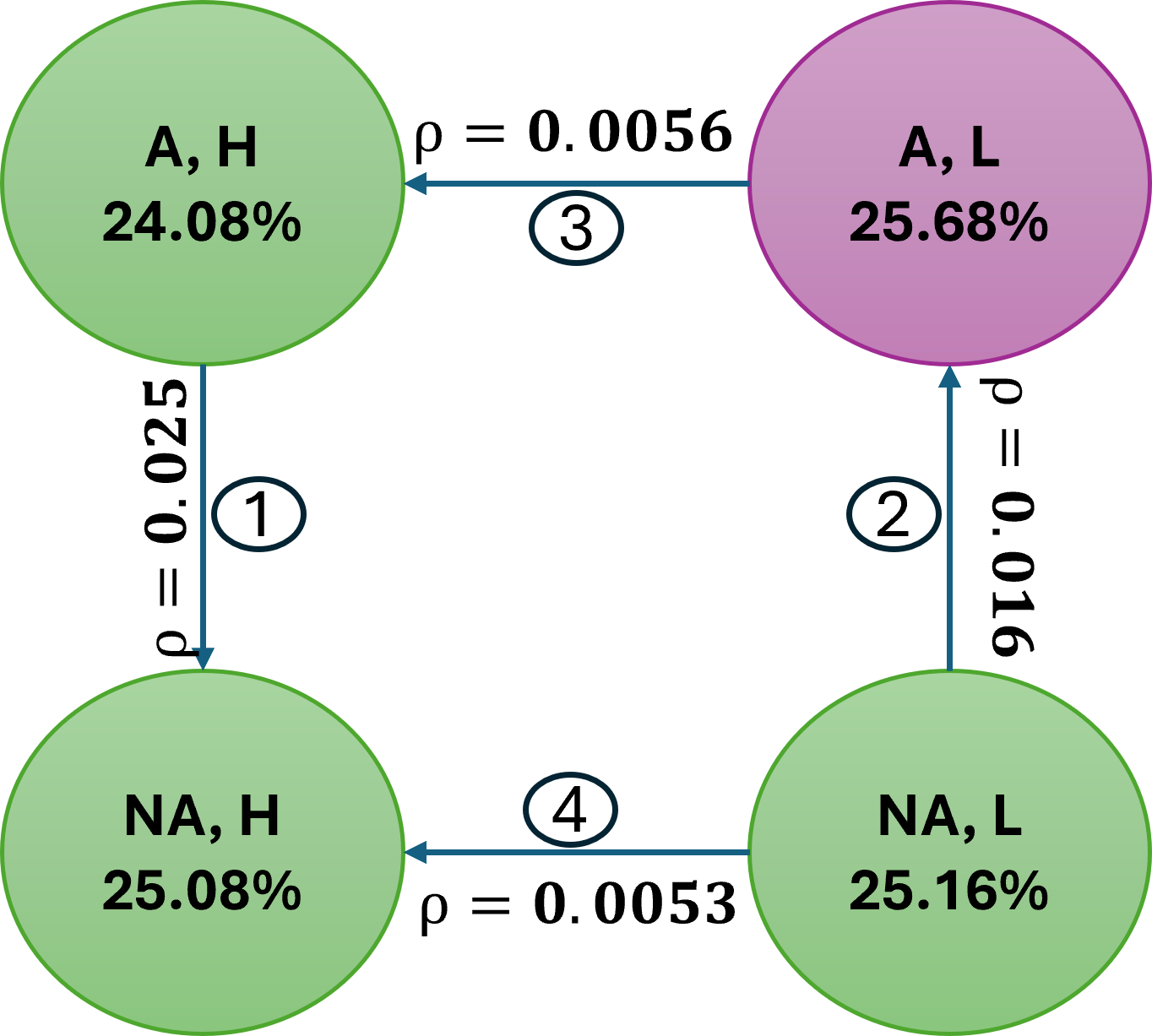}
    \caption{}
  \end{subfigure}
\caption{\textbf{Evolutionary dynamics without committed defenders ($z=0$).}
    (a) Stationary probabilities of the states $(A,H)$, $(A,L)$, $(NA,H)$, and $(NA,L)$ as a function of the selection intensity $\beta$ 
    for the baseline parameters 
    ($N=100$, $c_{ah}=0.85$, $b_{ah}=1.90$, $c_{al}=0.1$, $b_{al}=1.60$, 
    $p_{dh}=0.82$, $p_{dl}=0.75$, $B_H=0.75$, $B_L=0.55$, $C_H=0.41$, 
    $C_L=0.20$, $W_H=0.22$, $W_L=0.10$) and no committed players ($z=0$). 
    As $\beta$ increases, the population gradually shifts toward the $(A,L)$ state, 
    showing the dominance of low-effort defenders when imitation pressure strengthens.
    (b) Embedded 4-state Markov chain illustrating the transitions among the states for selection intensity $\beta=0.1$ with the same parameters. 
    Edges represent transition $\rho$ values, and node labels show the stationary probabilities ($\pi_i$) for each state with the purple colour for fixating strategy. The numbers in the oval shape show the risk-dominant strategy from Table \ref{tab:6}. Under the absence of committed players ($z=0$), the system converges primarily to the $(A,L)$ representing a risk-prone environment where attackers dominate.}
\label{fig:5}
\end{figure}

The finite-population analysis in Section \ref{3} reveals that cyber defence can stabilise only when defence is sufficiently effective and rewarding and when attacking becomes risky and costly, confirming that meaningful cyber resilience requires strong defence performance rather than simply penalising attackers. However, due to the high cost of $H$ defence, ordinary defenders naturally converge toward $L$ defence, leading to persistent attack success. To motivate the ordinary defenders, we introduce a fraction $z$ of committed defenders (zealots) who permanently adopt $H$ and examine whether they can shift behaviour in the remaining population. Figure~\ref{fig:5} shows that when $z=0$, the system settles into an equilibrium dominated by $L$ defence and high attack frequency, yielding an insecure cyber ecosystem. The Markov diagram confirms that the population spends most time around low–defence states. Thus, when no committed defenders are present, the system tends toward a high-risk equilibrium where attackers maintain strategic advantage by attacking the vulnerable defenders.

\begin{figure}[H]
\centering
\includegraphics[width=1\linewidth,height=0.65\textheight,keepaspectratio]{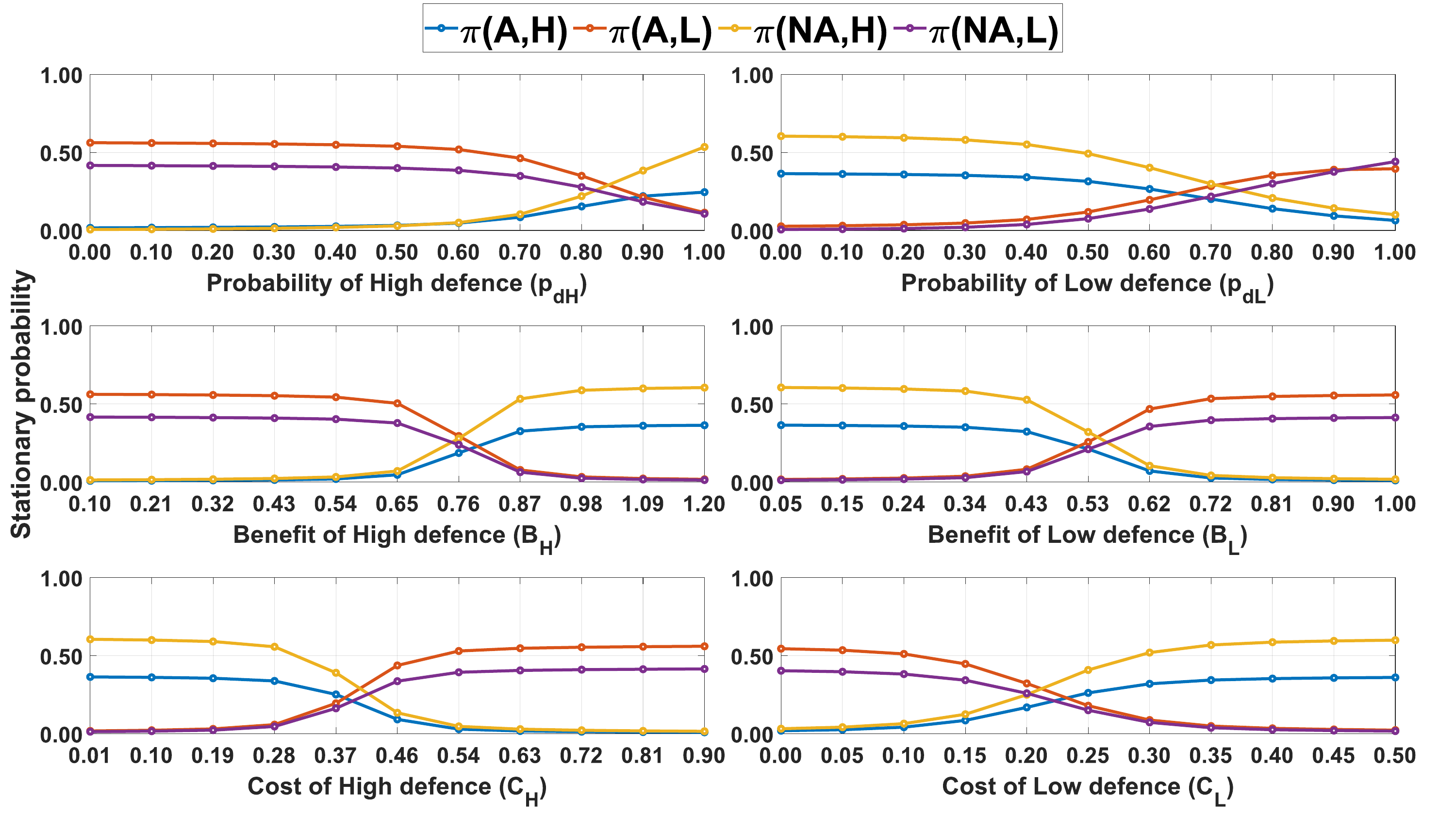}
\caption{Stationary strategy frequencies under parameter variations with no 
    committed defenders ($z = 0$, $\beta = 1$). While transitions between attack 
    and non-attack states are visible, low-effort defence $(NA,L)$ and attack 
    strategies $(A,L)$ retain significant probability mass across most parameter 
    ranges, indicating that high-effort defence $(NA,H)$ cannot dominate without 
    structural intervention.
    }
\label{fig:6}
\end{figure}

Figure~\ref{fig:6} shows the parameter sensitivity under the absence of committed defenders ($z = 0$) and moderate selection intensity ($\beta = 1$). Across all six parameters, $\pi(A,L)$ and 
$\pi(NA,L)$ retain significant probability mass, while $\pi(NA,H)$ remains low 
for most parameter ranges. Only under highly favourable conditions — high $p_{dH}$, 
high $B_H$, or low $C_H$ — does $\pi(NA,H)$ begin to rise, but it never dominates. This indicates that, when all defenders face the same high defence cost, evolutionary incentives strongly favour low-effort defence and persistent attacks. These baseline sweeps therefore establish a clear reference point showing that structural intervention is required to improve cyber resilience. This motivates the introduction of committed defenders ($z>0$) in the next section.

\begin{figure}[H]
\centering
\begin{subfigure}[t]{0.60\textwidth}
    \centering
    \includegraphics[width=\linewidth]{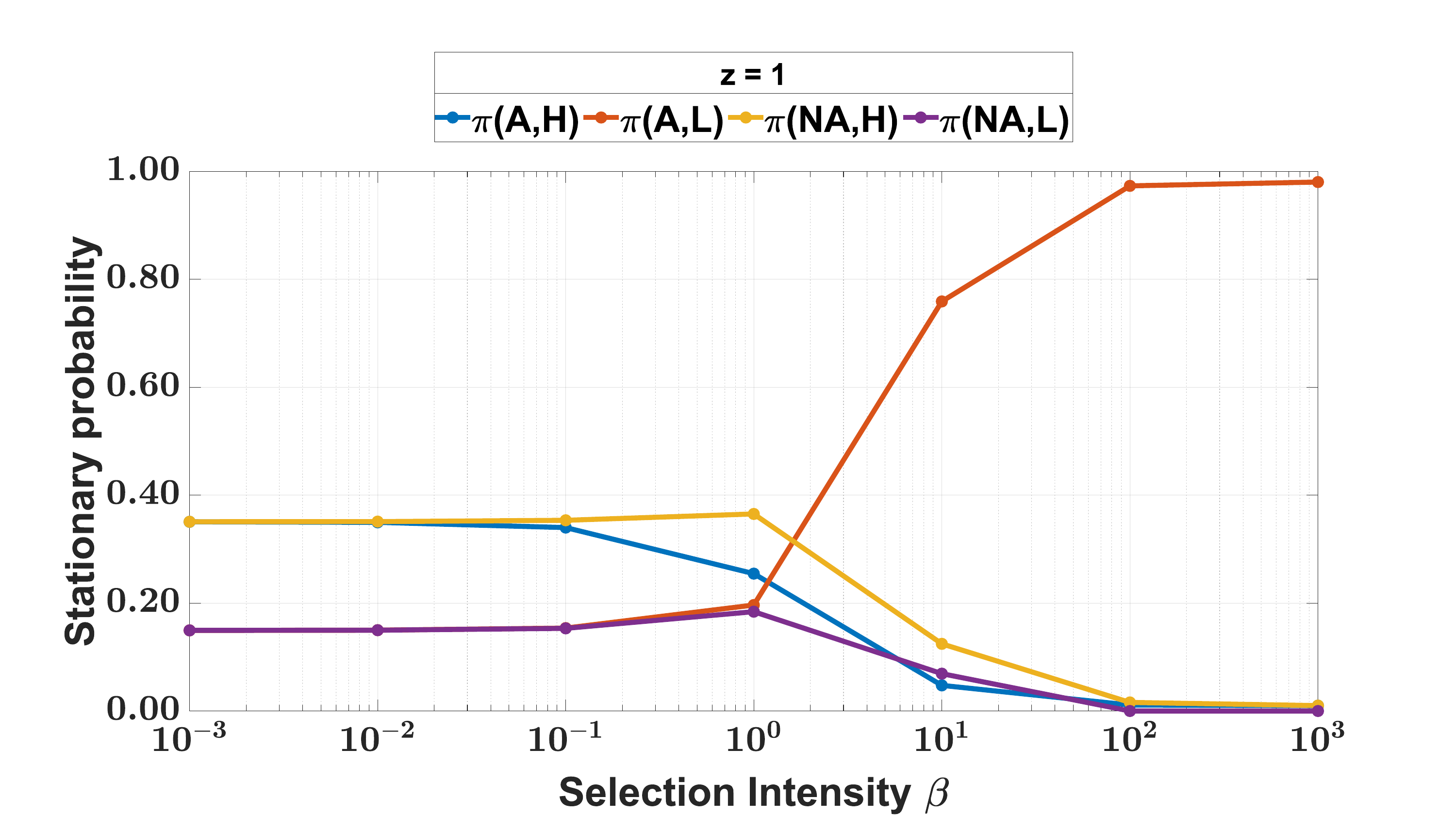}
    \caption{}
  \end{subfigure}
\begin{subfigure}[t]{0.38\textwidth}
    \centering
    \includegraphics[width=\linewidth]{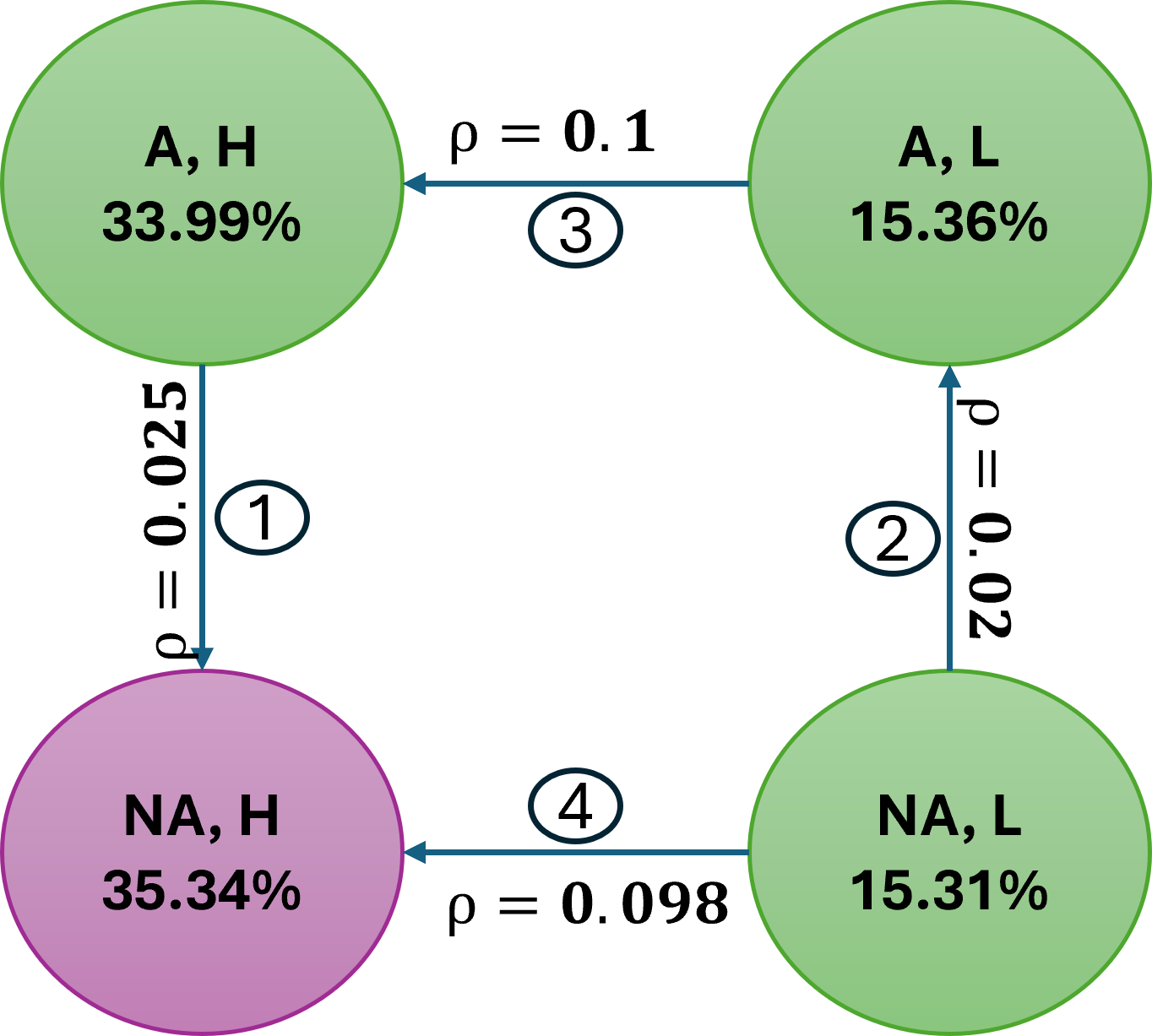}
    \caption{}
  \end{subfigure}
\caption{\textbf{Evolutionary dynamics with committed $H$ defenders ($z>0$)}. 
(a) Stationary probabilities of the four states $(A,H)$, $(A,L)$, $(NA,H)$, and $(NA,L)$ as a function of selection intensity $\beta$ under baseline parameters ($N=100$, $c_{ah}=0.85$, $b_{ah}=1.90$, $c_{al}=0.1$, $b_{al}=1.60$, $p_{dh}=0.82$, $p_{dl}=0.75$, $B_H=0.75$, $B_L=0.55$, $C_H=0.41$, $C_L=0.20$, $W_H=0.22$, $W_L=0.10$). Committed defenders motivate other defence players to adapt $H$ defence. For small commitment levels ($z=1$), the secure state $(NA,H)$ initially increases with selection intensity ($\beta$), indicating that committed defenders can promote high defence. However, as $\beta$ becomes large, $(NA,H)$ loses stability due to the high defence cost $C_H$, and the system shifts towards $(A,L)$. In contrast, for larger commitment levels ($z=1$), as shown in Figure \ref{fig:12} (Appendix), $(NA,H)$ becomes stable across high $\beta$, demonstrating that sufficient commitment is required to sustain secure outcomes. In (b), embedded Markov chain representation shows similar behavior for $\beta=0.1$ that committed defenders shift the dynamics towards $(NA,H)$.}
\label{fig:7}
\end{figure}

\begin{figure}[H]
\centering
\includegraphics[width=1\linewidth]{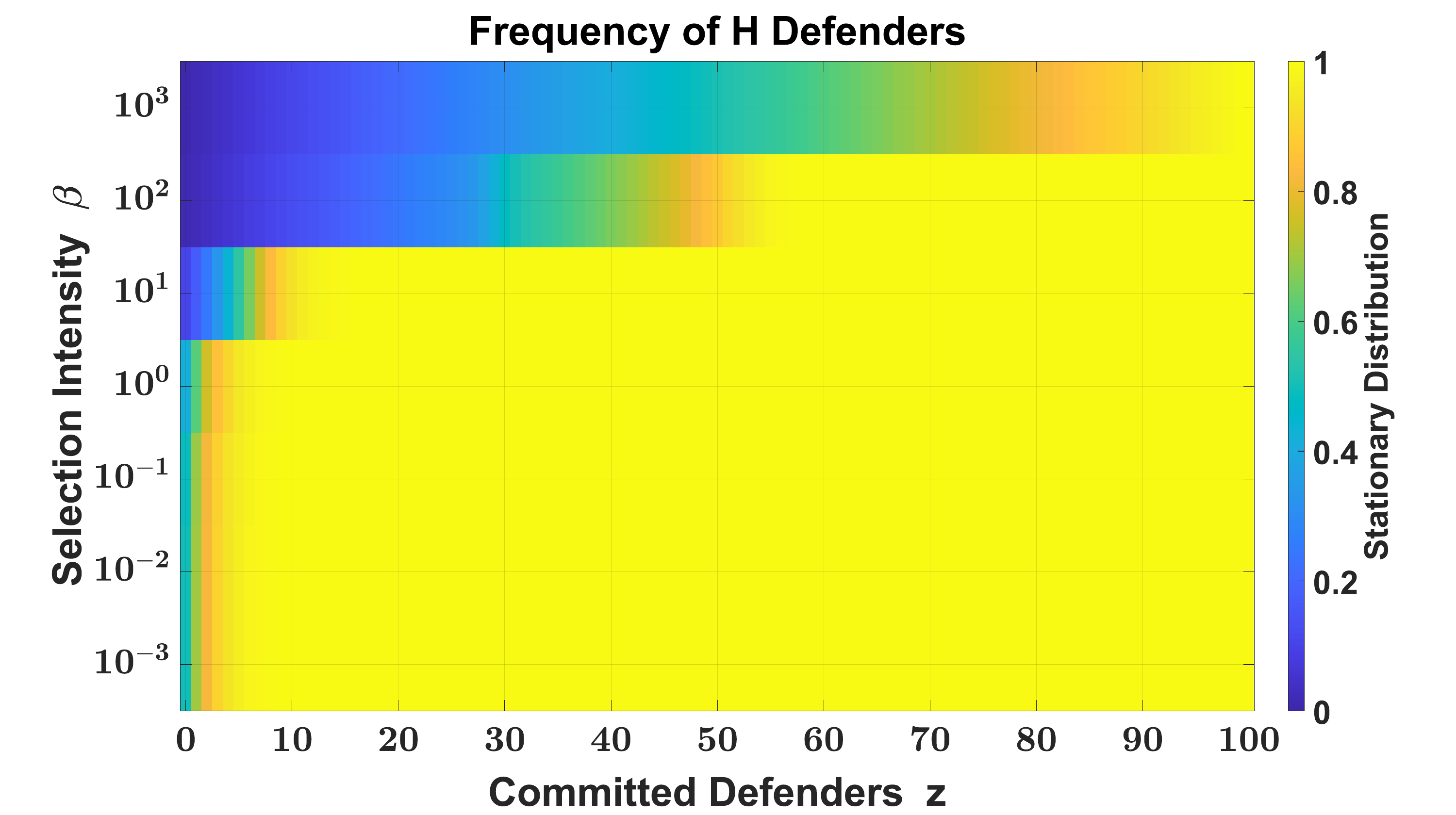}
\caption{Frequency of high-defence adoption $\pi_H = \pi(A,H) + \pi(NA,H)$ over the $(z,\beta)$ space. Even a small number of committed defenders ($z \geq 3$) is sufficient to drive widespread adoption of $H$ through imitation, particularly under moderate to strong selection. However, high adoption of $H$ does not necessarily imply convergence to the secure state $(NA,H)$, indicating a gap between defence adoption and true system stability. The same parameters are used as in Figure \ref{fig:7}.}
\label{fig:8}
\end{figure}

Introducing committed defenders consistently shifts the system towards high defence, as shown in Figures~\ref{fig:7} and \ref{fig:8}. The $\beta$-dependent dynamics (Figure~\ref{fig:7}) show that increasing the number of committed defenders $z$ raises the prevalence of $H$ defence, but stable convergence to the secure state $(NA,H)$ is achieved only when commitment is sufficiently high, as illustrated in Figure~\ref{fig:12} (Appendix). For small and moderate values of $z$, the system remains sensitive to the selection intensity $\beta$, and the population does not reliably settle in the secure equilibrium.

The heatmaps in Figures~\ref{fig:8} and \ref{fig:13} (Appendix) provide a complementary view by showing the overall distribution of strategies across parameter ranges. These results confirm that even a small number of committed defenders ($z \geq 3$) can significantly increase the adoption of $H$ defence. However, high adoption alone does not guarantee system-level security. Similar trend is observed in Figure~\ref{fig:14} in the Appendix, which shows that for a moderate level of commitment ($z=10$), the system remains highly sensitive to parameter changes. In particular, the cost of $H$ defence ($C_H$) plays a critical role, increasing $C_H$ quickly reduces the adoption of $(NA,H)$ and shifts the population towards attack-prone states. 

This highlights a key distinction between widespread adoption of high defence and true stability of the system. Introducing committed defenders increases the adoption of high defence through social learning. However, commitment alone does not stabilise the secure equilibrium, as high cost of defence $C_H$ limits full convergence to $(NA,H)$. Thus, while commitment shifts behaviour, it does not resolve the underlying economic constraint. This motivates the need for policy interventions that reduce the cost burden of $H$ defence. In the next section, we introduce a subsidy mechanism to address this limitation and stabilise secure system behaviour.

\section{Subsidised Committed Defenders}
\label{5}

To overcome the limitation of high defence cost $C_H$, discussed in the previous section, we introduce a policy mechanism in which committed defenders are subsidised so that they no longer bear the full cost of high defence. The use of subsidies to stabilise equilibria in non-cooperative games is formally studied in \cite{bilo2026compensate}, where subsidies compensate players deviating from a target profile. This reflects practical cybersecurity governance, where governments financially support critical infrastructure operators to strengthen resilience.

In the baseline case, both ordinary and committed high defenders pay cost $C_H$, and their payoffs against attacker $(A)$ or non-attacker $(NA)$ are:
\[
f_H^A = p_{dH} B_H - C_H - (1 - p_{dH}) W_H, \qquad
f_H^{NA} = B_H - C_H.
\]

\noindent With subsidy, committed defenders pay no cost, and their payoff becomes:
\[
f_H^A(z) = p_{dH} B_H - (1 - p_{dH}) W_H = f_H^A + C_H,
\qquad
f_H^{NA}(z) = B_H = f_H^{NA} + C_H.
\]

\noindent Because strategy imitation depends on payoffs, the effective value of playing $H$ increases with the number of subsidised defenders $z$. When $z$ out of $N$ defenders are subsidised, the subsidy-adjusted average payoff for strategy $H$ is:
\[
f_H^A(z) = f_H^A + \frac{z}{N} C_H,
\qquad
f_H^{NA}(z) = f_H^{NA} + \frac{z}{N} C_H.
\]

Low defenders retain their original payoffs $f_L^A$ and $f_L^{NA}$, and thus the Fermi imitation update rule becomes:
\[
p_{L \to H}^{(A)}(z)
  = \frac{1}{1+\exp\!\left(-\beta \left[f_H^A(z) - f_L^A\right]\right)},
\qquad
p_{H \to L}^{(A)}(z)
  = \frac{1}{1+\exp\!\left(-\beta \left[f_L^A - f_H^A(z)\right]\right)},
\]
and similarly when facing a non-attacker using $f_H^{NA}(z)$ and
$f_L^{NA}(z)$.
This subsidy mechanism increases the perceived payoff advantage of $H$ defenders, making $H$ defence more attractive and accelerating convergence to the secure equilibrium $(NA,H)$.

\begin{figure}[H]
\begin{subfigure}[t]{0.60\textwidth}
    \centering
    \includegraphics[width=1\linewidth]{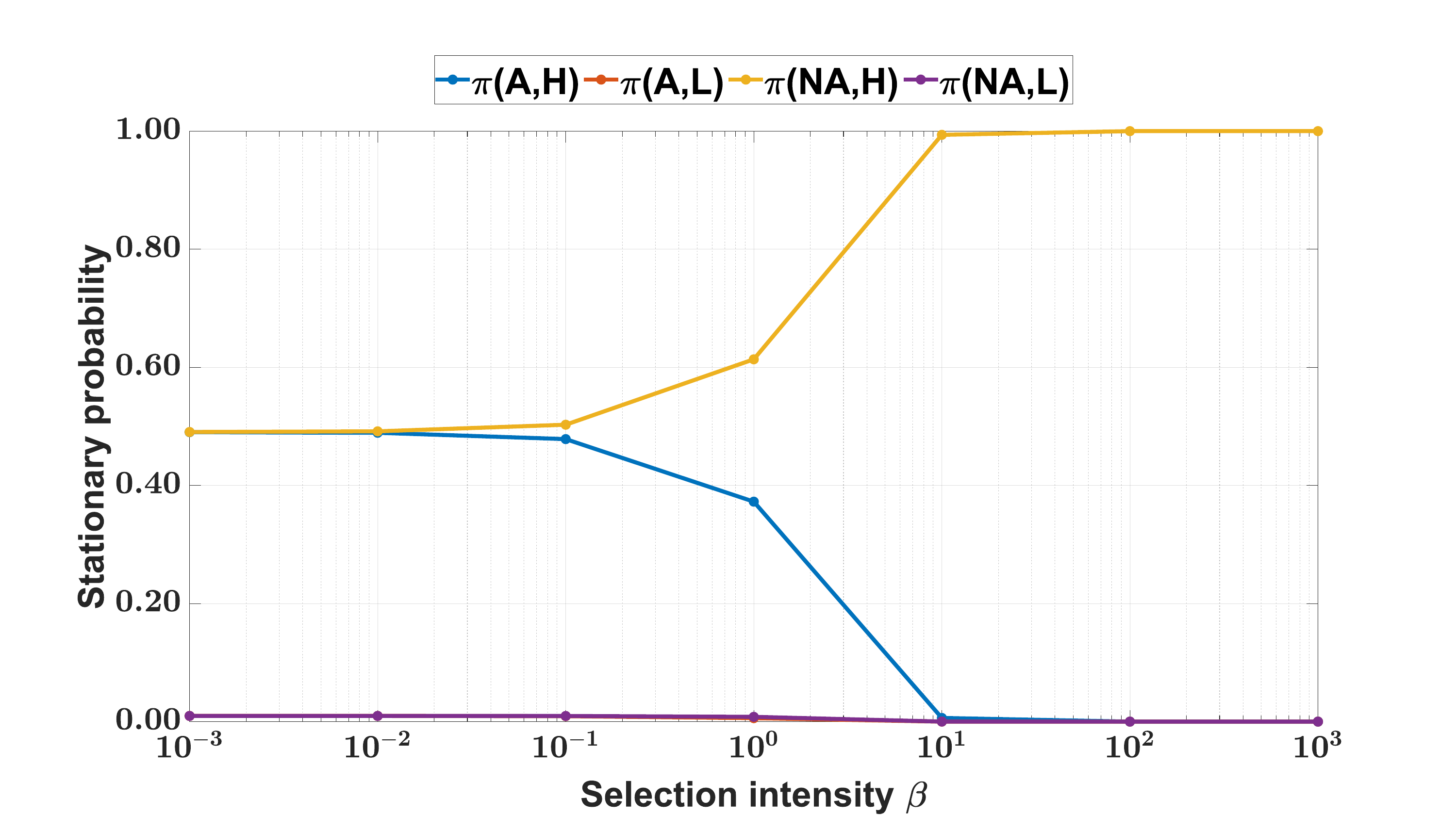}
    \caption{}
  \end{subfigure}
\begin{subfigure}[t]{0.38\textwidth}
    \centering
    \includegraphics[width=1\linewidth]{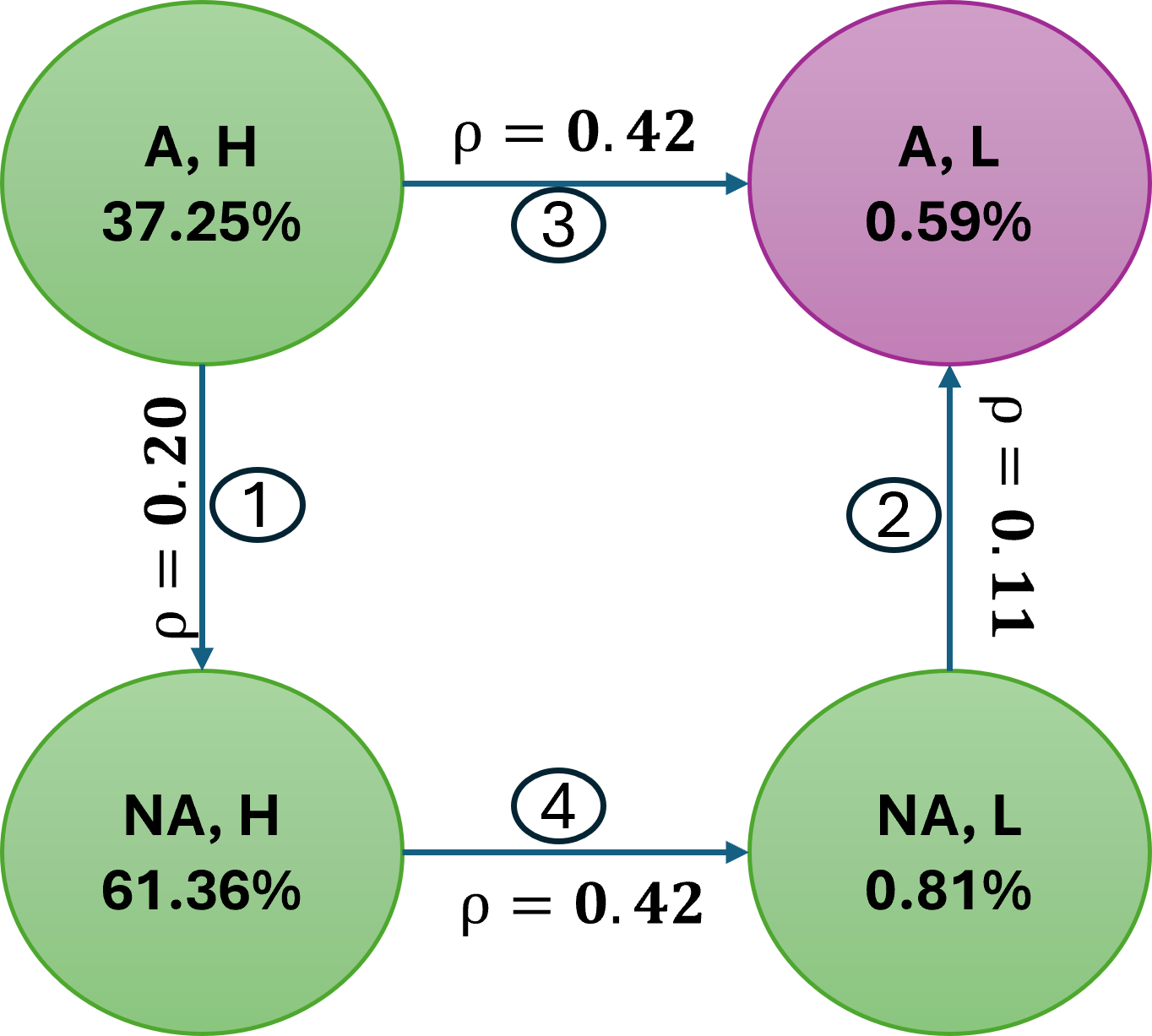}
    \caption{}
  \end{subfigure}
\caption{\textbf{Stationary outcomes under subsidised committed defenders.}
\textbf{(a)} Stationary probabilities of the four strategy states across selection intensity $\beta$ when committed defenders receive subsidy. The secure state $(NA,H)$ starts dominating from $z=6$, indicating strong stabilisation of high defence under subsidy. 
\textbf{(b)} Corresponding Markov transition diagram showing transition probabilities $\rho$ on edges and stationary state probabilities at nodes. The numbers in the oval shape show the risk-dominant strategy from Table \ref{tab:6}. The system resides in $(NA,H)$, while attacker-dominant states become rare, demonstrating that subsidy shifts the population towards a safe equilibrium and suppresses transitions toward vulnerable states. Baseline parameters used are $N=100$, $c_{ah}=0.85$, $b_{ah}=1.90$, $c_{al}=0.10$, $b_{al}=1.60$, 
    $p_{dh}=0.82$, $p_{dl}=0.75$, $B_H=0.75$, $B_L=0.55$, $C_H=0.41$, 
    $C_L=0.20$, $W_H=0.22$, $W_L=0.10$, $z=10$, $\beta=0.1$.}
\label{fig:9}
\end{figure}

Introducing subsidies for committed $H$ defenders eliminates the cost disadvantage of $H$ defenders and increases their payoff advantage over $L$ defenders. This shifts the imitation dynamics and $H$ defence becomes more attractive, encouraging ordinary defenders to adopt $H$ strategies through social learning. As shown in Figures~\ref{fig:9} \ref{fig:15} (in the Appendix), subsidy fundamentally changes the evolutionary landscape for very less number of subsidised committed defenders. The heatmaps in Figure~\ref{fig:16} in the appendix provide a complementary view by showing the overall distribution of strategies across varying levels of $\beta$. The stationary distribution concentrates on the secure state $(NA,H)$, while attacker-dominant states collapse to near zero. Even without subsidy, a small number of committed defenders ($z>0$) can shift the population toward the high-defence state $(NA,H)$, but full stability is achieved only at $z=N$ (Figure~\ref{fig:7}), whereas with subsidy the same shift occurs for $z>0$ and stable convergence to $(NA,H)$ is attained at much lower commitment levels (e.g., $z \approx 6$ for $N=100$). The subsidy converts safety improvements into real economic benefit and suppresses incentives for attack, leading to a more resilient cyber ecosystem. 

\paragraph{Subsidy Threshold and Optimal Allocation.}

Prior work on subsidy design shows that governments choose subsidy levels subject to budget constraints \cite{sadana2025subsidizing,li2024government}. In our model, subsidies increase the payoff advantage of high defence, making $H$-defence more attractive and accelerating convergence toward the secure state $(NA,H)$. Even a small number of subsidised defenders improves the adoption of $H$ defence and becomes more pronounced as the number of subsidised defenders increases. However, the benefits of the subsidy do not increase indefinitely. Instead, there exists a threshold level of commitment beyond which additional subsidy yields diminishing improvements in system behaviour. In practice, subsidy policies are constrained by limited resources. Given a government budget $G$, the feasible level of subsidised defenders satisfies $z \leq G/C_H$. These results indicate that effective cyber defence does not require subsidising all defenders. Rather, a moderate and well-targeted level of subsidy is sufficient to promote high-defence adoption and stabilise secure outcomes without excessive public expenditure. While prior work models environment dynamics through stochastic game transitions \cite{luo2025evolutionary}, we instead consider structural interventions that modify payoff incentives directly. This allows us to capture how economic constraints and policy support influence the emergence of secure behaviour in cyber systems. To verify that these insights are not driven by specific parameter choices, we next examine the robustness of the model under a wide range of randomly generated environments.

\begin{figure}[H]
\centering
    \includegraphics[width=0.85\textwidth]{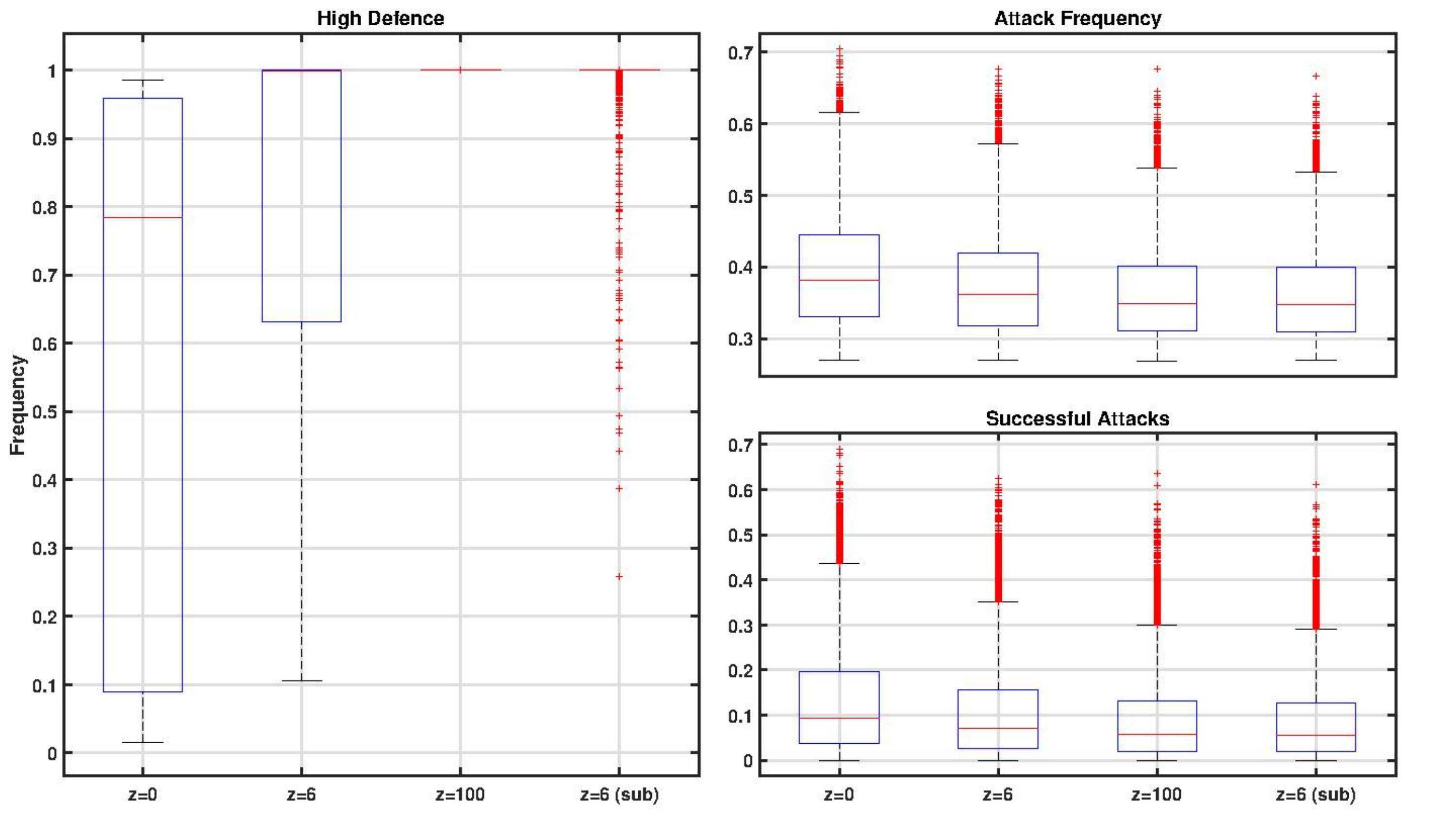}
\caption{\textbf{Distribution of outcomes across 10,000 random games ($\beta=1$, $z \in \{0,6,100\}$).}
The left panel shows the distribution of high-defence frequency, while the top-right and bottom-right panels show attack frequency and successful attack rates, respectively. Increasing the number of committed defenders shifts the population toward high defence and reduces successful attacks. While partial commitment ($z=6$) improves average outcomes, variability remains high. In contrast, subsidised commitment at $z=6$ achieves near-complete adoption of high defence and further concentrates the distribution of successful attacks at lower values, matching the behaviour of full commitment ($z=100$). This shows that subsidy enables stable and reliable security at significantly lower levels of commitment.}
\label{fig:10}
\end{figure}

\paragraph{Random Games Analysis:}
We simulated 10,000 randomly generated games  under uniformly sampled parameters while preserving the logical constraints ($C_L < C_H$, $B_L < B_H$, $W_L < W_H$, $p_{dL} < p_{dH}$, $b_{aL} < b_{aH}$, $c_{aL} < c_{aH}$, and $z \in [0,10]$), for robustness checking of our general findings \cite{duong2025random,bashir2026co}. For each parameter configuration, we computed the stationary distribution and recorded the resulting defence and attack frequencies, as summarised in Table~\ref{tab:7} and illustrated in Figure \ref{fig:10}. To better reflect the actual security outcome, we also measured the rate of \emph{successful attacks}. For each random game, a successful attack is defined as the probability that an attack occurs and the defence fails. This is computed from the stationary distribution by weighting each attack state with its corresponding defence failure probability. Formally, the successful attack rate is given by
\begin{equation}
\pi_{\text{succ}} = \pi_{(A,H)}(1 - p_{dH}) + \pi_{(A,L)}(1 - p_{dL}),
\end{equation}
where $\pi_{(A,H)}$ and $\pi_{(A,L)}$ are the stationary probabilities of the attack states under high and low defence, respectively. To highlight the deterrence mechanism, we analyse a filtered set of games satisfying $f_A^{H} < f_A^{L}$ to ensure that attacking against high defence is meaningfully less attractive. Under this condition, the effect of commitment becomes clearer: committed defenders drive the population toward high-defence states and reduce the rate of \emph{successful attacks}. The distribution of these successful attack rates across random games is shown in the boxplots in Figure~\ref{fig:10} and Table \ref{tab:7}.

\begin{table}[H]
\centering
\caption{Summary statistics over 10,000 randomly generated games for $\beta = 1$. Results are reported as mean $\pm$ standard deviation.}
\label{tab:7}
\begin{tabular}{lccc}
\toprule
\textbf{Case} & \textbf{Attack attempts} & \textbf{High defence} & \textbf{Successful attacks} \\
\midrule
$z=0$            & $0.393 \pm 0.077$ & $0.583 \pm 0.397$ & $0.134 \pm 0.124$ \\
$z=6$           & $0.375 \pm 0.072$ & $0.806 \pm 0.298$ & $0.108 \pm 0.109$ \\
$z=100$          & $0.362 \pm 0.064$ & $\mathbf{1.000 \pm 0.000}$ & $0.091 \pm 0.094$ \\
$z=6$ (subsidy) & $\mathbf{0.360 \pm 0.064}$ & $0.998 \pm 0.024$ & $\mathbf{0.088 \pm 0.092}$ \\
\bottomrule
\end{tabular}
\end{table}
\noindent The results in Table~\ref{tab:7} demonstrate that increasing the number of committed defenders has a clear impact on system security. Moving from $z=0$ to $z=6$, high-defence adoption increases substantially (from $0.583$ to $0.806$), while successful attacks decrease from $0.134$ to $0.108$, indicating improved resilience. Full commitment ($z=100$) leads to complete adoption of high defence and further reduces successful attacks to $0.091$. Introducing subsidy significantly strengthens this effect at low levels of commitment. With only $z=6$ subsidised committed defenders, high-defence adoption becomes nearly complete ($0.998$), matching the behaviour observed under full commitment, while successful attacks further decrease to $0.088$. Compared to the baseline case ($z=0$), this corresponds to an approximate $34\%$ reduction in successful attacks. These results highlight that while commitment improves average system performance, subsidy enables near-complete adoption of high defence at much lower commitment levels, effectively approaching the benefits of full commitment without requiring all defenders to be committed.

This indicates that although commitment does not eliminate attack attempts, it significantly reduces the effectiveness of attacks and subsidy further reduce the required number of committed defenders. Subsidising committed defenders removes the cost disadvantage of high defence and fundamentally changes the evolutionary dynamics. 
Subsidy enables stable convergence to the secure state even with a small number of committed defenders, reduces successful attacks, and improves defender incentives. 
We compare three scenarios. 
In the baseline case, high defence cost leads to weak defence adoption and persistent attacks. 
With committed defenders, adoption of high defence increases, but the system remains unstable due to the cost burden. 
In contrast, subsidy removes this economic barrier, enabling stable convergence to the secure equilibrium with fewer committed players. 
This comparison highlights that incentive design is essential for achieving both stability and resilience. While the subsidy mechanism stabilises the secure equilibrium $(NA,H)$ and reduces the prevalence of attacker–dominant states, it is important to evaluate its broader system-level impact. Therefore, to assess the trade-offs between security and economic outcomes, we next analyse the resulting social welfare under different levels of commitment and subsidy.

\subsection{Social Welfare and Collective Security Impact}

\begin{figure}[H]
\centering
\begin{subfigure}[t]{0.49\textwidth}
    \centering
    \includegraphics[width=\linewidth]{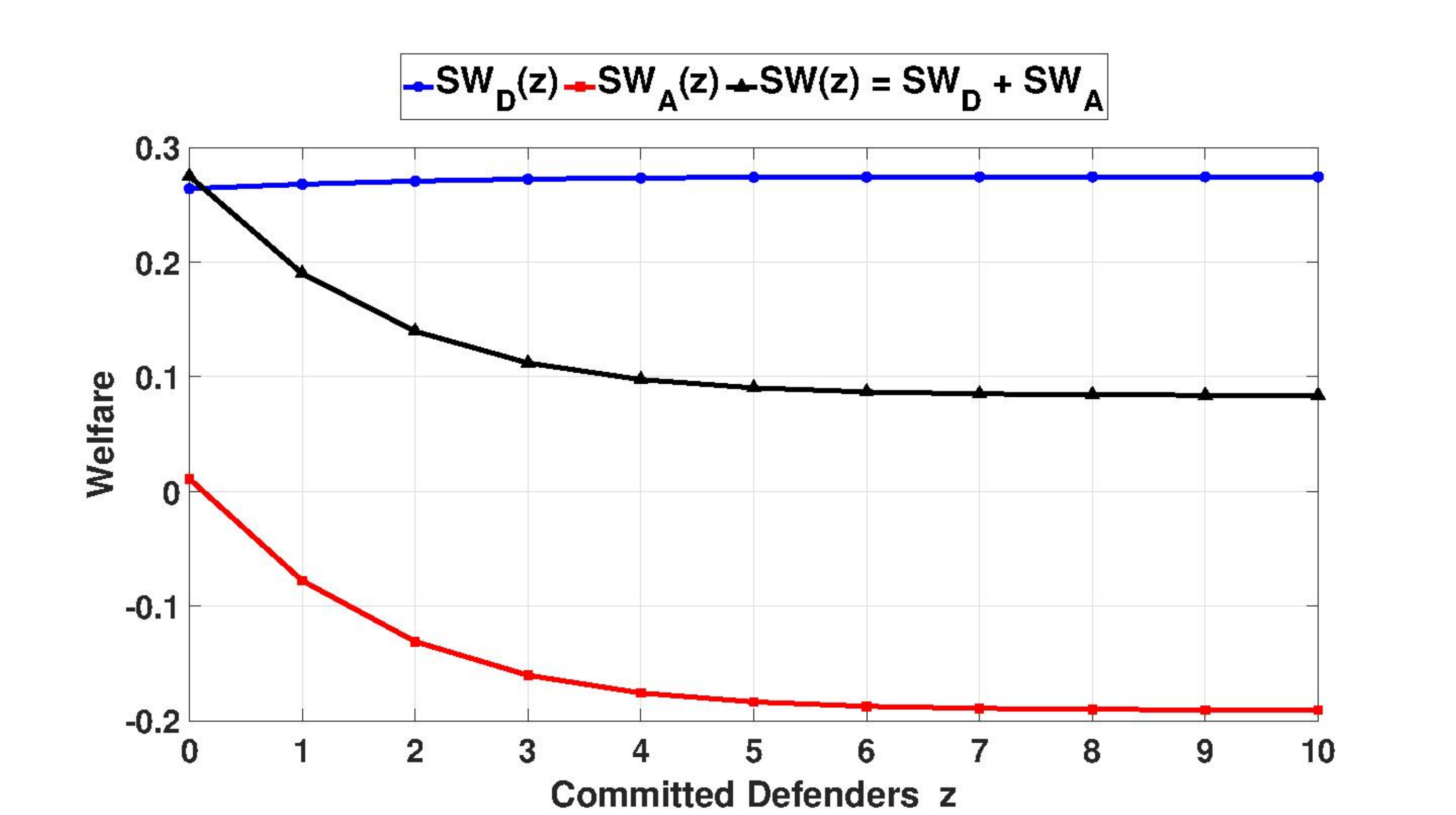}
    \caption{}
  \end{subfigure}
  \begin{subfigure}[t]{0.49\textwidth}
    \centering
    \includegraphics[width=\linewidth]{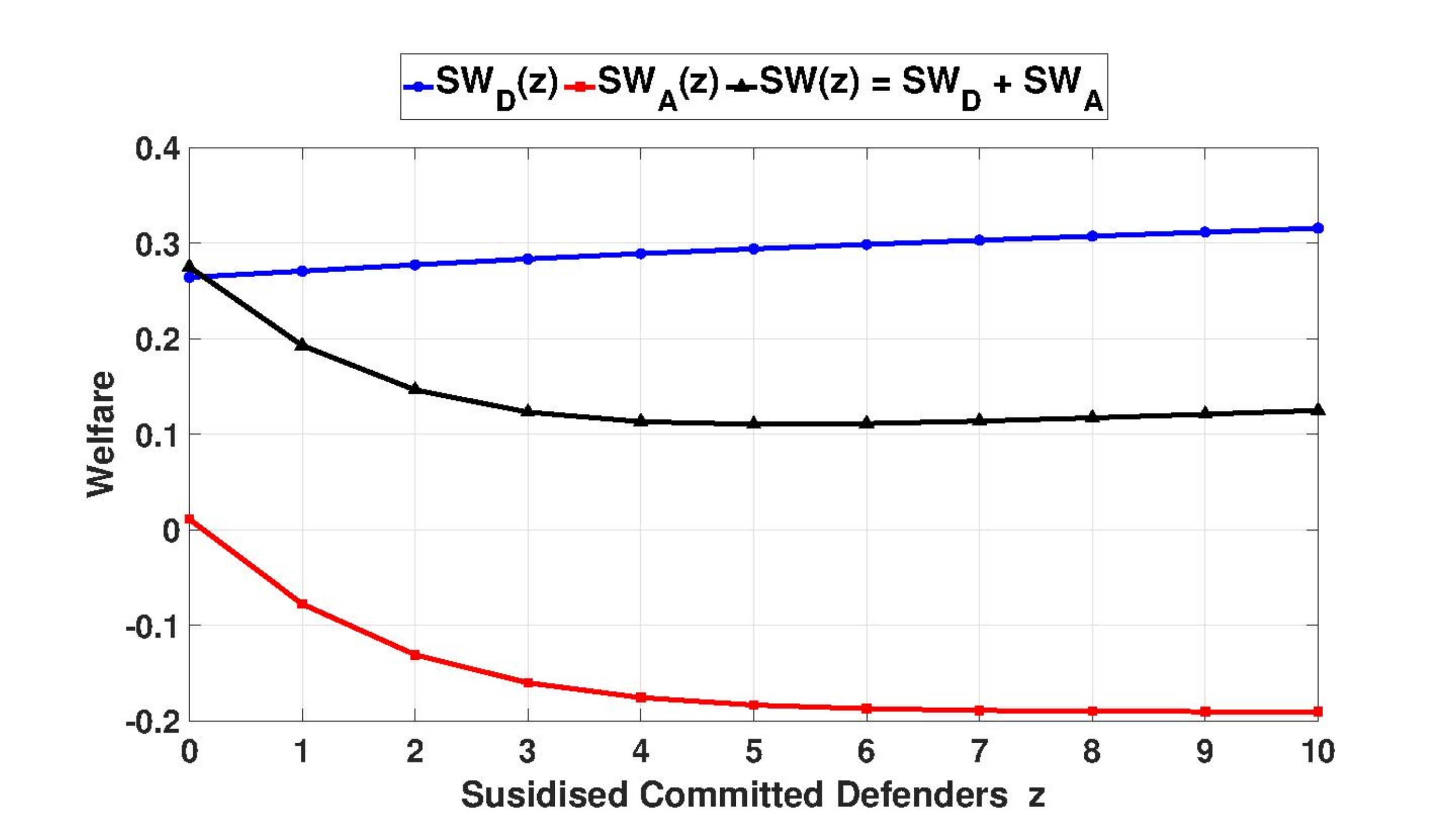}
    \caption{}
  \end{subfigure}
\caption{Effect of Committed Defenders on Social Welfare.
Sub figure (a) shows that as the number of committed defenders z increases, defender welfare $SW_D(z)$ (blue) stays nearly flat while attacker welfare $SW_A(z)$ (red) drops sharply, and their sum $SW(z)$ (black) also drops. Subfigure (b) adds the case with subsidised committed defenders z, where attacker welfare (red) drops more steeply and defender welfare (blue) rises slightly with z, indicating that subsidising committed defenders benefits defenders. Overall, more committed defenders hurt attackers but do not meaningfully benefit defenders unless the cost of $H$ defence is reduced. Baseline parameters are used for these plots ($N=100$, $c_{ah}=0.85$, $b_{ah}=1.90$, $c_{al}=0.1$, $b_{al}=1.60$, 
    $p_{dh}=0.82$, $p_{dl}=0.75$, $B_H=0.75$, $B_L=0.55$, $C_H=0.41$, 
    $C_L=0.20$, $W_H=0.22$, $W_L=0.10$).}
\label{fig:11}
\end{figure}

This mechanism also resolves the earlier limitation of differential access. Previously, committed defenders improved system safety but did not increase defender welfare, as in Figure~\ref{fig:11} (a). Under subsidisation, defender social welfare rises with $z$, attacker welfare declines (Figure~\ref{fig:11} (b)). To evaluate the broader consequences of differential AI access on collective security, we analyse the \emph{social welfare} of the system, defined as the long–run expected utility at evolutionary equilibrium. Let $\mathcal{S}=\{(A,H),(A,L),(NA,H),(NA,L)$ be the set of macro–states, and let $\pi_s(z,\beta)$ denote the stationary probability of state $s\in\mathcal{S}$ under $(z,\beta)$. If $f_d(s)$ and $f_a(s)$ are defender and attacker payoffs in state $s$, respectively, then following~\cite{han2024evolutionary}, total social welfare is computed as:
\begin{equation}
\label{eq:SW_total}
SW(z,\beta)=\sum_{s\in\mathcal{S}}\pi_s(z,\beta)\,[f_d(s)+f_a(s)].
\tag{19}
\end{equation}
This quantity reflects the total expected benefit across the full population when the system has settled into its stationary distribution. Using baseline parameter values, per–state social welfare is the sum of defender and attacker payoffs:
\[
\begin{aligned}
Payoff(A,H)&=f_A^H+f_H^A,\quad &Payoff(A,L)&=f_A^L+ f_L^A,\\
Payoff(NA,H)&=f_{NA}^H+ f_H^{NA},\quad &Payoff(NA,L)&=f_{NA}^L+ f_L^{NA}.
\end{aligned}
\]

\noindent Since total $SW$ aggregates payoffs of both defenders and attackers, reductions in attacker payoff may decrease total welfare even when security improves. Therefore, from a cybersecurity perspective, it is more informative to evaluate defender and attacker welfare separately:
\begin{equation}
\begin{split}
    SW_D(z,\beta)=\sum_{s\in\mathcal{S}}\pi_s(z,\beta)\,f_D(s), \quad
SW_A(z,\beta)=\sum_{s\in\mathcal{S}}\pi_s(z,\beta)\,f_A(s).
\end{split}
\label{eq:SW_def}
\tag{20}
\end{equation}

\begin{table}[h]
\centering
\caption{Impact of committed defenders and subsidy on social welfare.}
\label{tab:social_welfare}
\begin{tabular}{lccc}
\toprule
\textbf{Case} & $\mathbf{SW_D}$ & $\mathbf{SW_A}$ & $\mathbf{SW}$ \\
\midrule
$z=0$ (no subsidy)    & 0.264 & 0.011  & 0.275 \\
$z=10$ (no subsidy)   & 0.274 & -0.190 & 0.084 \\
$z=10$ (with subsidy) & \textbf{0.315} & \textbf{-0.191} & \textbf{0.125} \\
\bottomrule
\end{tabular}
\end{table}
Subfigure~\ref{fig:11}(a) shows that increasing $z$ sharply reduces attacker social welfare, confirming reduced attacker advantage—a desirable security outcome. However, defender welfare remains nearly constant because defenders continue to pay the high cost $C_H$. In contrast, when subsidies are introduced (Subfigure~\ref{fig:11}(b)), defender welfare increases with $z$ while attacker welfare declines further, indicating that subsidy not only enhances security but also improves defender incentives. 

Overall, these results demonstrate that while committed defenders alone improve system safety, subsidy is essential to translate these improvements into tangible welfare gains and stable adoption of high defence. Having established both the security and welfare benefits of subsidy-enhanced differential access, we now discuss the broader policy implications, practical relevance, and limitations of this mechanism.

\section{Discussion}
\label{7}

This study shows that cyber defence effectiveness in finite populations is strongly influenced by empowering a small number of committed defenders with high–capability AI defence tools. While traditional cybersecurity strategies often emphasise increasing penalties or disruption costs for attackers~\cite{aggarwal2015cyber,khadam2023punish}, our findings show that strengthening defence incentives has a stronger and more stable impact at the population level. In contrast to behaviour-based defence mechanisms \cite{hu2025security}, our approach targets structural incentives through commitment and subsidy to shift population-level dynamics. 

Introducing committed $H$ defenders reshapes the evolutionary dynamics and creates a critical-mass phenomenon that a small number ($z \geq 1$ in $N = 100$) is sufficient to shift the system toward secure outcomes, even when attackers remain present. However, commitment alone is not sufficient. Without subsidy, the high cost $C_H$ limits sustained adoption of $H$ defence and prevents improvements in defender welfare. The introduction of subsidy resolves this limitation. By reducing the cost burden for committed defenders, subsidy accelerates convergence to $(NA,H)$, increases $SW_D$, and further suppresses attacker incentives. This highlights a key insight: stability and welfare improvement require both commitment and incentive support, which is in line with recent work on differential AI access~\cite{ee2025asymmetry} and decentralised cyber defence models~\cite{liu2022modeling}. They support practical policy mechanisms such as selective subsidy, enterprise-tier access control, and trusted defender certification. Overall, the study confirms that strengthening defence incentives can stabilise cyber ecosystems more effectively than penalising attackers. Differential AI access together with a small group of subsidised committed defenders can accelerate convergence to secure equilibria and improve defender welfare by reducing attack incentives.

\begin{tcolorbox}[colback=gray!10, colframe=black, title=\textbf{Policy Implications for Secure AI Deployment}]
\begin{itemize}
    \item \textbf{Targeted defence investment may outperform uniform allocation.} 
    The model suggests that a small number of subsidised, committed high-defence actors ($z > 1$ for $N=100$) can drive system-wide adoption of strong defence.
    \item \textbf{Subsidising defenders is more effective than penalising attackers.} 
    The results indicate that improving defence incentives leads to more stable and resilient outcomes than increasing attacker costs.
    \item \textbf{Differential AI access can support practical governance.} 
    This can be interpreted as controlled access to high-capability AI defence tools improving security without requiring full population deployment.
    \item \textbf{Subsidy can improve both security and welfare.} 
    The model shows that reducing defence costs for committed defenders increases adoption of $H$ strategies and raises defender welfare.
    \item \textbf{Certain actors may function as committed defenders.} 
    This can be interpreted as security agencies, regulated vendors, or trusted organisations acting as stabilising nodes that promote cooperative defence behaviour.
    
\end{itemize}
\end{tcolorbox}

These results provide a clear foundation for designing AI-enabled cybersecurity policies, targeted resource allocation strategies and strategic AI governance in national cybersecurity infrastructures. 
While this study provides a rigorous EGT analysis of cyber defence in finite populations, several limitations remain. Attackers in our model are homogeneous and do not adapt over time. In practice, attackers learn, coordinate, and evolve strategies, which may change system dynamics. Extending the model to adaptive attackers is an important direction for future work. In addition, the subsidy mechanism is modelled in a simplified way. Real-world implementation would involve budget constraints, regulatory design, and institutional factors that are not captured here. These aspects require further empirical and policy-driven analysis.

\paragraph{Future Work}

While the present model captures the essential interplay between differential AI access, committed defence, and subsidised incentives, several important extensions remain open. First, future work should incorporate richer forms of strategic heterogeneity, including adaptive attackers, defenders with diverse resource levels, and endogenous investment decisions. Such extensions would bring the model closer to real cyber environments, where both offensive and defensive capabilities evolve over time and actors differ substantially in their ability to bear the cost of advanced protection.

Second, it would be valuable to examine broader classes of institutional incentives. In addition to subsidies, future models could study mixed policy instruments such as rewards for adopting strong defence, penalties for weak defence, and adaptive regulatory mechanisms that respond to observed attack prevalence. Analysing such hybrid incentive schemes could clarify how public policy and organisational governance jointly shape defensive behaviour across interconnected sectors \cite{liu2025evolution}.

Third, a particularly important direction is the design of principled subsidy mechanisms under explicit budget constraints. Rather than treating subsidy as exogenous, future research could investigate how limited public or institutional resources should be allocated across defenders, which agents should be prioritised, and under what conditions targeted intervention yields the greatest systemic benefit. Addressing these questions would help translate the present theoretical results into actionable guidance for cybersecurity policy and AI governance.

Overall, these directions would further connect evolutionary game-theoretic modelling with deployable AI-enabled cybersecurity systems and deepen our understanding of how strategic incentives can be used to build resilient digital infrastructures under technological asymmetry.

\section{Conclusions} 
\label{8}

In this paper, we studied cyber attack-defence dynamics in finite populations through an evolutionary game-theoretic framework and investigated how differential access to AI-enabled defence capabilities shapes long-run security outcomes. Our results show that costly high-quality defence creates a structural obstacle to cyber resilience: even when powerful defensive tools are available, defenders are often driven toward cheaper and weaker strategies, allowing attacks to persist.

We then examined whether this vulnerability can be mitigated through differential AI access, committed defence, and targeted subsidies. Allowing some defenders to access stronger AI-based protection improves the defensive landscape, but this alone is not sufficient to secure the system. A small committed group of high-defence players can redirect the social-learning dynamics and increase the visibility of strong defence, yet commitment by itself does not reliably stabilise the secure equilibrium $(NA,H)$. The reason is fundamental: as long as high defence remains costly, most defenders still face incentives to avoid it.

The main contribution of the paper is to show that subsidised commitment overcomes this barrier. By removing the cost disadvantage for committed defenders, subsidies make strong defence more evolutionarily sustainable and amplify its influence on the wider population. Across extensive random game experiments, subsidised commitment consistently increases adoption of high defence, lowers successful attack rates, and accelerates convergence toward secure outcomes. The welfare analysis further strengthens this conclusion: commitment alone does not guarantee higher defender welfare, whereas subsidy-supported commitment improves defender outcomes while continuing to suppress attacker incentives.

Taken together, these findings show that stabilising cyber ecosystems depends less on punishing attackers than on strengthening the incentives and capacity of defenders. A relatively small but well-supported set of trusted defenders can generate system-wide security gains without requiring universal deployment of expensive technologies. This insight is especially relevant in AI-enabled cybersecurity, where capability asymmetries and resource constraints are likely to become increasingly important. More broadly, our results provide a theoretical foundation for policy interventions based on differential access, committed defence, and targeted financial support as practical tools for improving resilience in adversarial digital environments.

\section*{Acknowledgements}
T.A.H. and Z.S. are supported by EPSRC (grant EP/Y00857X/1). 
M.P. was supported by the Slovenian Research and Innovation Agency
(Grant No. P1-0403).

\bibliographystyle{unsrt}
\bibliography{Finite-bib}

\newpage
\setcounter{figure}{0}
\setcounter{equation}{0}
\setcounter{table}{0}
\renewcommand*{\thefigure}{A\arabic{figure}}
\renewcommand*{\theequation}{A\arabic{equation}}
\renewcommand*{\thetable}{A\arabic{table}}
\section*{Appendix}

This appendix provides additional technical details and supporting results for the main analysis. 
We first present the average payoffs used to derive the evolutionary dynamics for both attacker and defender populations in Section \ref{3}. We then provide the full Markov transition matrix $M$ that governs the stochastic evolution between the four strategy pairs in Section \ref{4}. 

The remaining figures offer supplementary numerical results, including extended analyses of commitment levels and parameter sensitivity for committed players supporting Figures \ref{fig:7} and \ref{fig:8}, and the impact of subsidy supporting Figure \ref{fig:9} in the main text. These results support the main findings by illustrating how system behaviour changes across different parameter regimes and confirming the robustness of the secure equilibrium under varying conditions.

\noindent \textbf{Attacker-side Payoffs:}\\
When defender population is all defence $(D)$:
\begin{equation}
    f_A^D = \frac{(-c_a+b_a(1-p_d))N}{N}, \qquad
    f_{NA}^D = \frac{(0)N}{N}
\end{equation}
When defender population is all not defence $(ND)$:
\begin{equation}
    f_A^{ND} = \frac{(-c_a+b_a)N}{N}, \qquad
    f_{NA}^{ND} = \frac{(0)N}{N}
\end{equation}
\textbf{Defender-side Payoffs:}\\
When attacker population is all attack $(A)$:
\begin{equation}
    f_D^A = \frac{(-c_d+p_db_d-w(1-p_d))N}{N}, \qquad
    f_{ND}^A = \frac{(-w)N}{N}
\end{equation}
When attacker population is all not attack $(NA)$:
\begin{equation}
    f_D^{NA} = \frac{(-c_d+b_d)N}{N}, \qquad
    f_{ND}^{NA} = \frac{(0)N}{N}
\end{equation}
\textbf{Markov transition matrix $M$:}
\renewcommand{\arraystretch}{2.5}
\begin{equation}
\scriptsize
M =
\begin{pmatrix}
1 - \tfrac{1}{2}(\rho_{A \to NA}^{D} + \rho_{D \to ND}^{A}) 
& \tfrac{1}{2}\rho_{A \to NA}^{D} 
& \tfrac{1}{2}\rho_{D \to ND}^{A} 
& 0 \\

\tfrac{1}{2}\rho_{NA \to A}^{D} 
& 1 - \tfrac{1}{2}(\rho_{NA \to A}^{D} + \rho_{D \to ND}^{NA}) 
& 0 
& \tfrac{1}{2}\rho_{D \to ND}^{NA} \\

\tfrac{1}{2}\rho_{ND \to D}^{A} 
& 0 
& 1 - \tfrac{1}{2}(\rho_{ND \to D}^{A} + \rho_{A \to NA}^{ND}) 
& \tfrac{1}{2}\rho_{A \to NA}^{ND} \\

0 
& \tfrac{1}{2}\rho_{ND \to D}^{NA} 
& \tfrac{1}{2}\rho_{NA \to A}^{ND} 
& 1 - \tfrac{1}{2}(\rho_{ND \to D}^{NA} + \rho_{NA \to A}^{ND})
\end{pmatrix}
\end{equation}

\begin{figure}[H]
\centering    
    \includegraphics[width=1\linewidth]{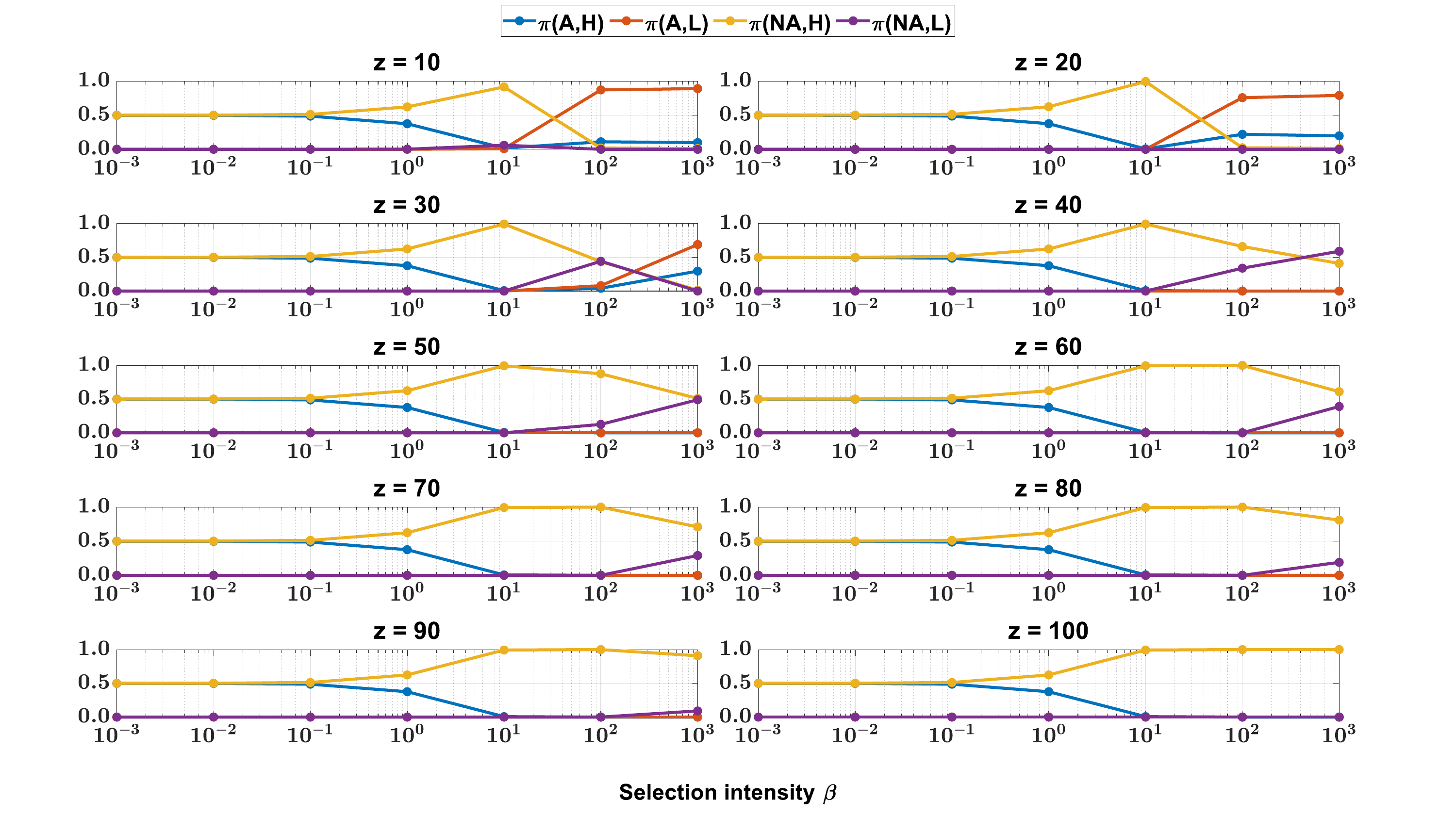}
\caption{Figure shows the stationary distribution across different levels of commitment $z$. For small and moderate values of $z$, the system remains sensitive to the selection intensity $\beta$, and the secure state $(NA,H)$ is not fully stabilised. As $z$ increases, the system gradually transitions towards stability. Only when $z$ approaches the total population size does the stationary distribution concentrate entirely on $(NA,H)$, indicating that full commitment is required to guarantee robustness in the absence of subsidy. Baseline parameters used are $N=100$, $c_{ah}=0.85$, $b_{ah}=1.90$, $c_{al}=0.10$, $b_{al}=1.60$, 
    $p_{dh}=0.82$, $p_{dl}=0.75$, $B_H=0.75$, $B_L=0.55$, $C_H=0.41$, 
    $C_L=0.20$, $W_H=0.22$, $W_L=0.10$, $z=10$, $\beta=0.1$.}
\label{fig:12}
\end{figure}
\begin{figure}[H]
\centering
\includegraphics[width=1\linewidth]{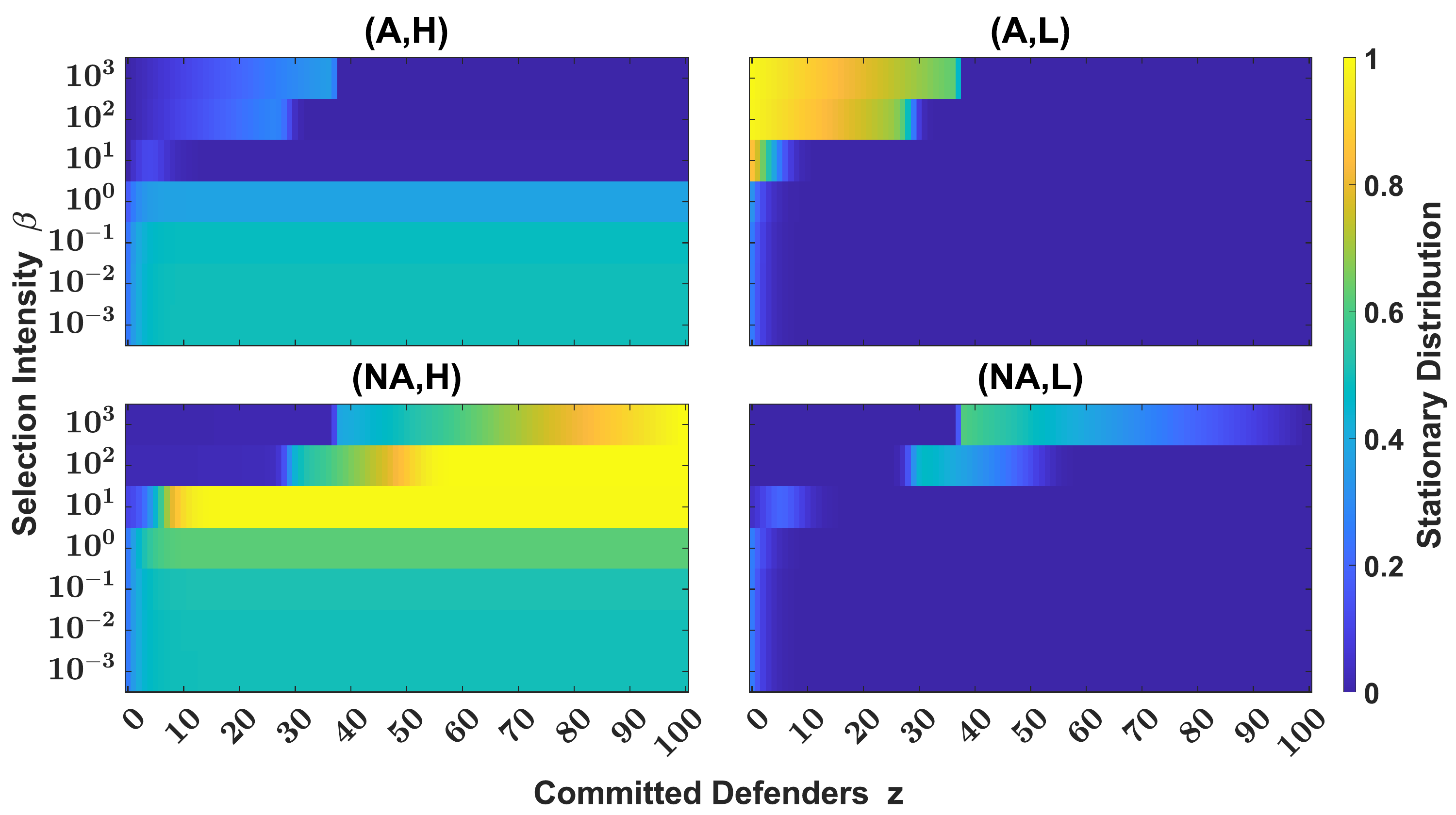}
\caption{Stationary distributions of the four states $(A,H)$, $(A,L)$, $(NA,H)$, and $(NA,L)$ over the $(z,\beta)$ space. Increasing the number of committed defenders $z$ shifts the population towards high-defence states, reducing the attacker-dominated state $(A,L)$. However, the secure equilibrium $(NA,H)$ becomes dominant only at higher levels of commitment and stronger selection intensity $\beta$. This highlights that while committed defenders promote high defence, the high cost $C_H$ limits full stabilisation of the secure state when commitment is low. Baseline parameters are used for these plots ($N=100$, $c_{ah}=0.85$, $b_{ah}=1.90$, $c_{al}=0.1$, $b_{al}=1.60$, $p_{dh}=0.82$, $p_{dl}=0.75$, $B_H=0.75$, $B_L=0.55$, $C_H=0.41$, $C_L=0.20$, $W_H=0.22$, $W_L=0.10$).}
\label{fig:13}
\end{figure}

\begin{figure}[H]
\centering
\includegraphics[width=1.03\linewidth,height=0.65\textheight,keepaspectratio]{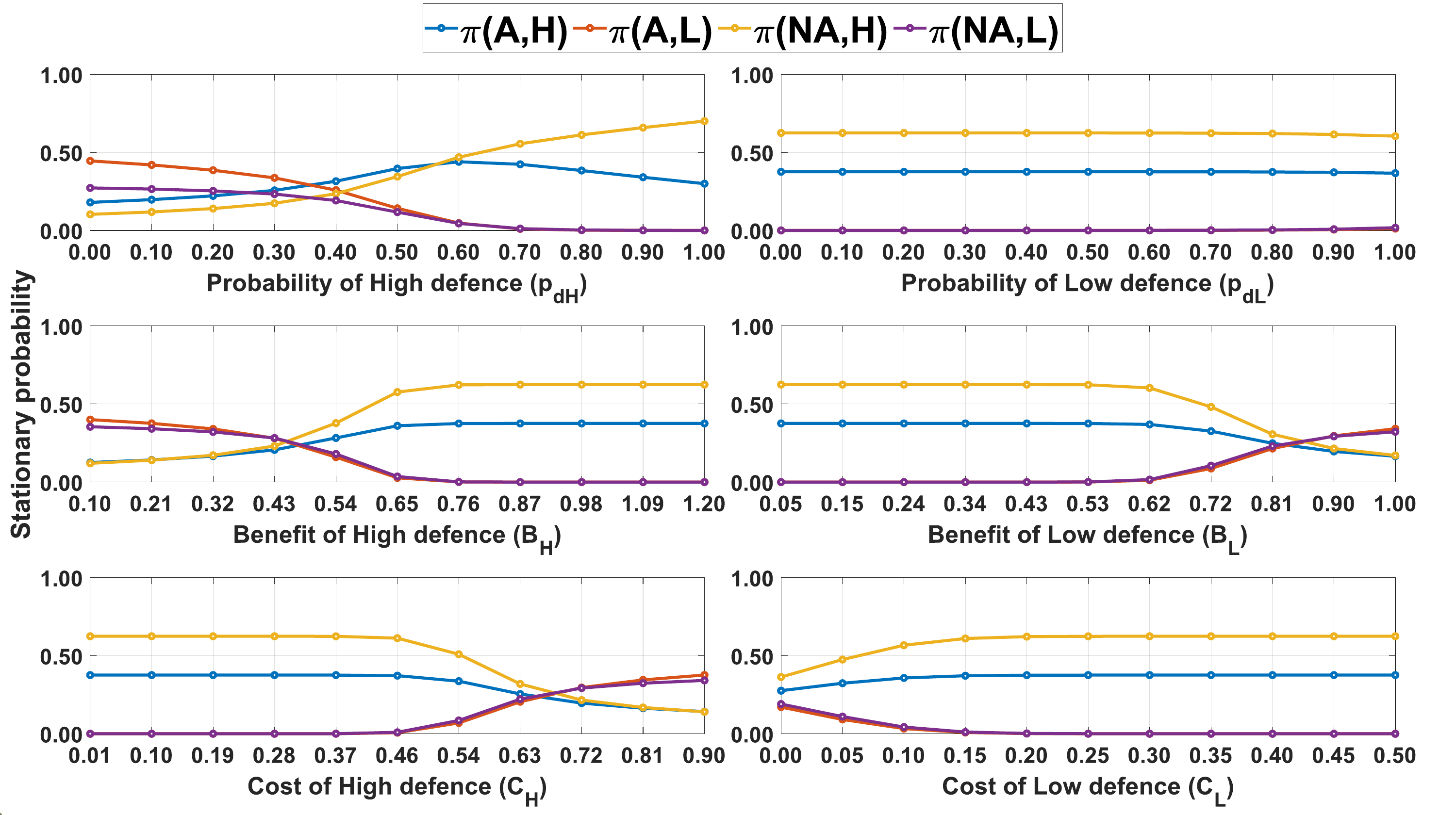}
\caption{\textbf{Parameter sensitivity analysis for $z=10$.} Stationary probabilities of $(A,H)$, $(A,L)$, $(NA,H)$, and $(NA,L)$ are shown as key model parameters vary. Increasing the probability and benefit of high defence ($p_{dH}, B_H$) promotes the secure state $(NA,H)$, while increasing the cost of high defence ($C_H$) reduces its adoption and shifts the population towards attack states. Variations in low-defence parameters ($p_{dL}, B_L, C_L$) have a weaker or mixed effect. Overall, even with committed defenders ($z=10$) and ($\beta=1$), the system remains sensitive to parameter changes, and high defence is not consistently stabilised due to its cost.}
\label{fig:14}
\end{figure}

\begin{figure}[H]
\centering    
    \includegraphics[width=1\linewidth]{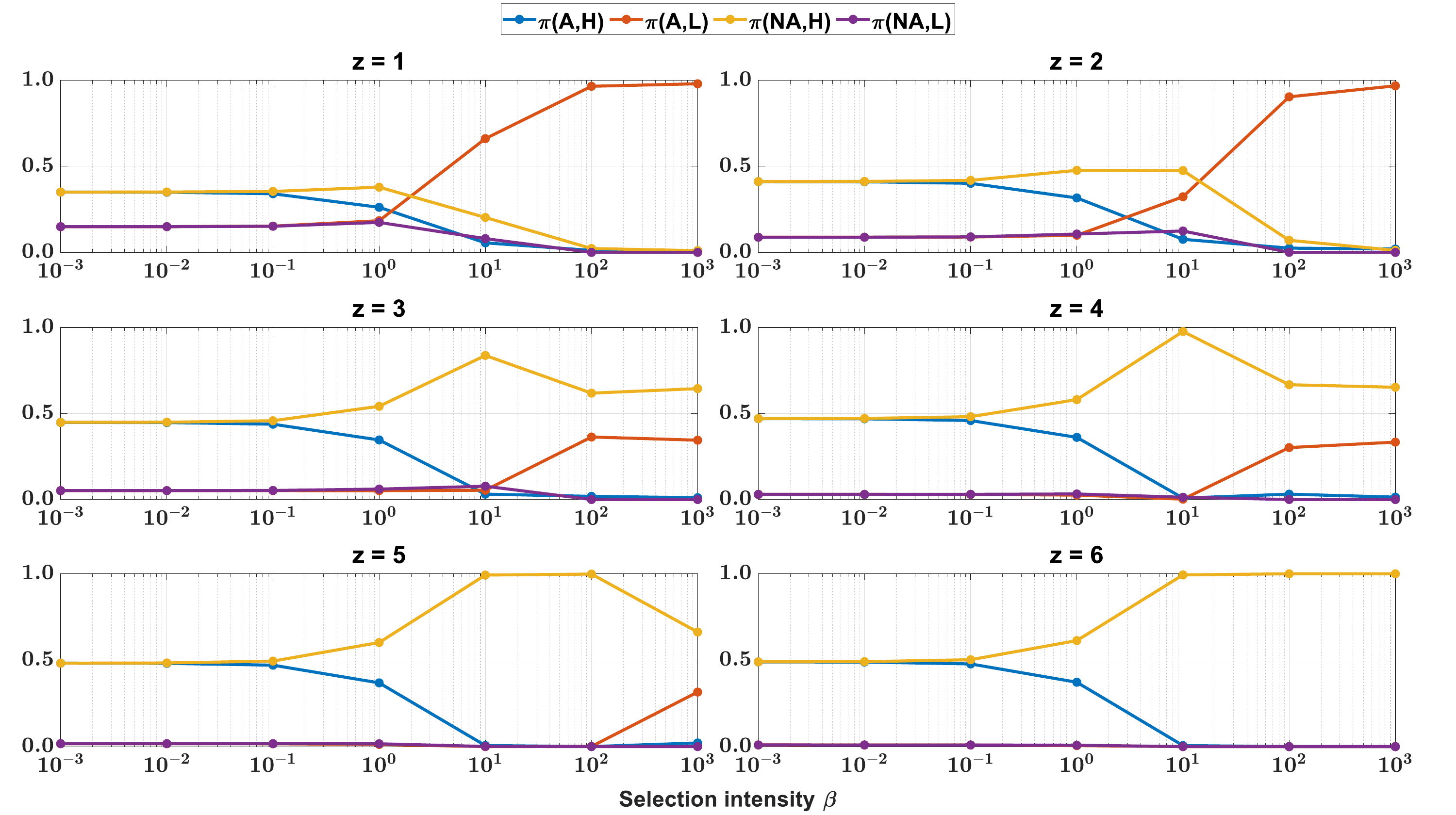}
\caption{Figure shows the stationary distribution across different levels of commitment $z$ with subsidy. For small and moderate values of $z$, the system remains sensitive to the selection intensity $\beta$, and the secure state $(NA,H)$ is not fully stabilised. As $z$ increases, the system gradually transitions towards stability. Only when $z$ approaches the total population size does the stationary distribution concentrate entirely on $(NA,H)$, indicating that full commitment is required to guarantee robustness in the absence of subsidy. Baseline parameters used are $N=100$, $c_{ah}=0.85$, $b_{ah}=1.90$, $c_{al}=0.10$, $b_{al}=1.60$, 
    $p_{dh}=0.82$, $p_{dl}=0.75$, $B_H=0.75$, $B_L=0.55$, $C_H=0.41$, 
    $C_L=0.20$, $W_H=0.22$, $W_L=0.10$, $z=10$, $\beta=0.1$.}
\label{fig:15}
\end{figure}

Committed defenders play a dominant stabilising role in the finite population dynamics, as shown in Figure \ref{fig:9} and \ref{fig:10}.
With strong commitment ($z=10$), the population concentrates around the safe equilibrium $(\textit{NA},H)$ and becomes largely insensitive to parameter variations, indicating high systemic robustness.

\begin{figure}[H]
\centering
\begin{subfigure}[t]{\textwidth}
\centering
\includegraphics[width=0.85\linewidth]{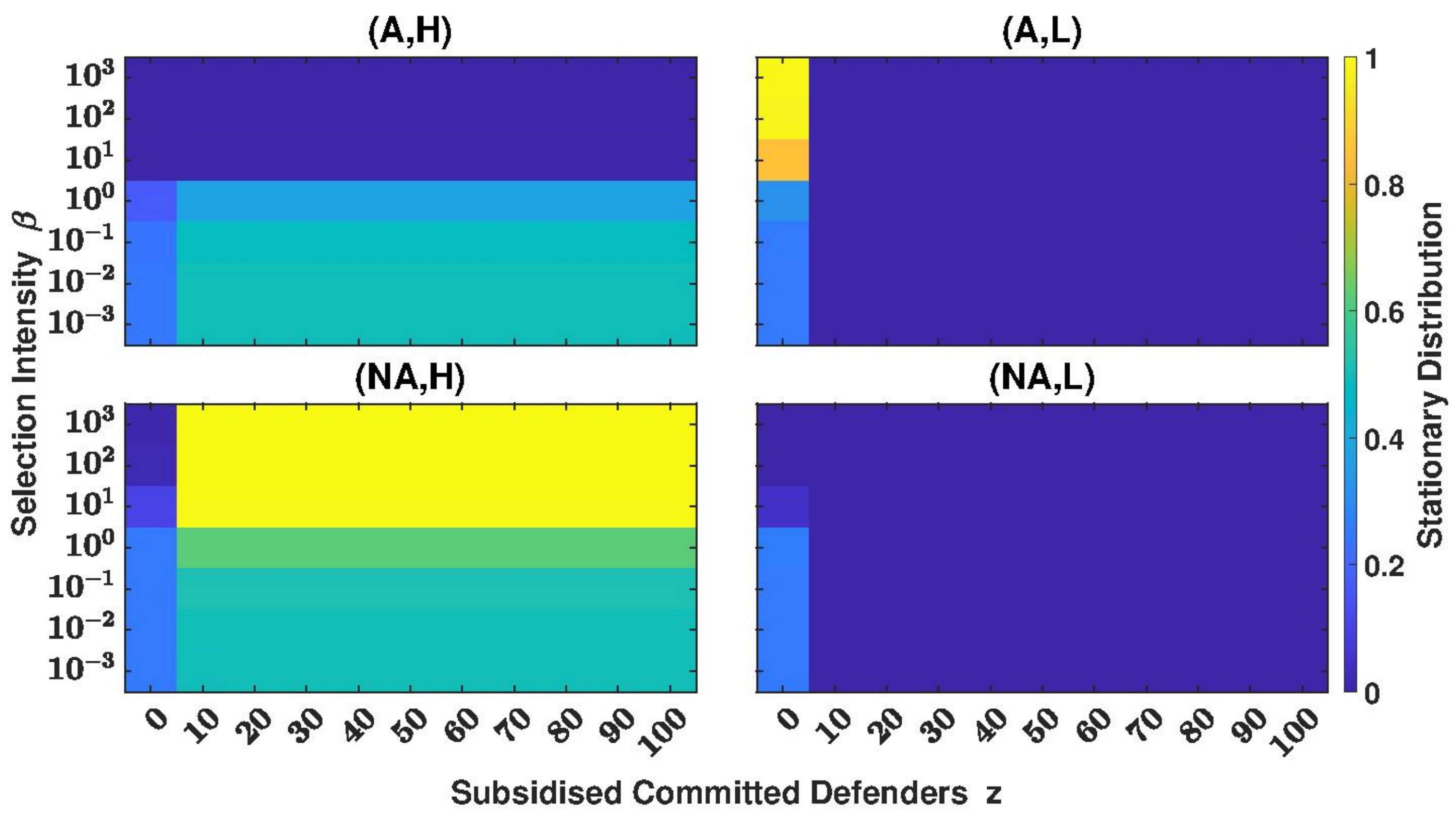}
\caption{}
\end{subfigure}
\begin{subfigure}[t]{\textwidth}
    \centering
    \includegraphics[width=0.85\linewidth]{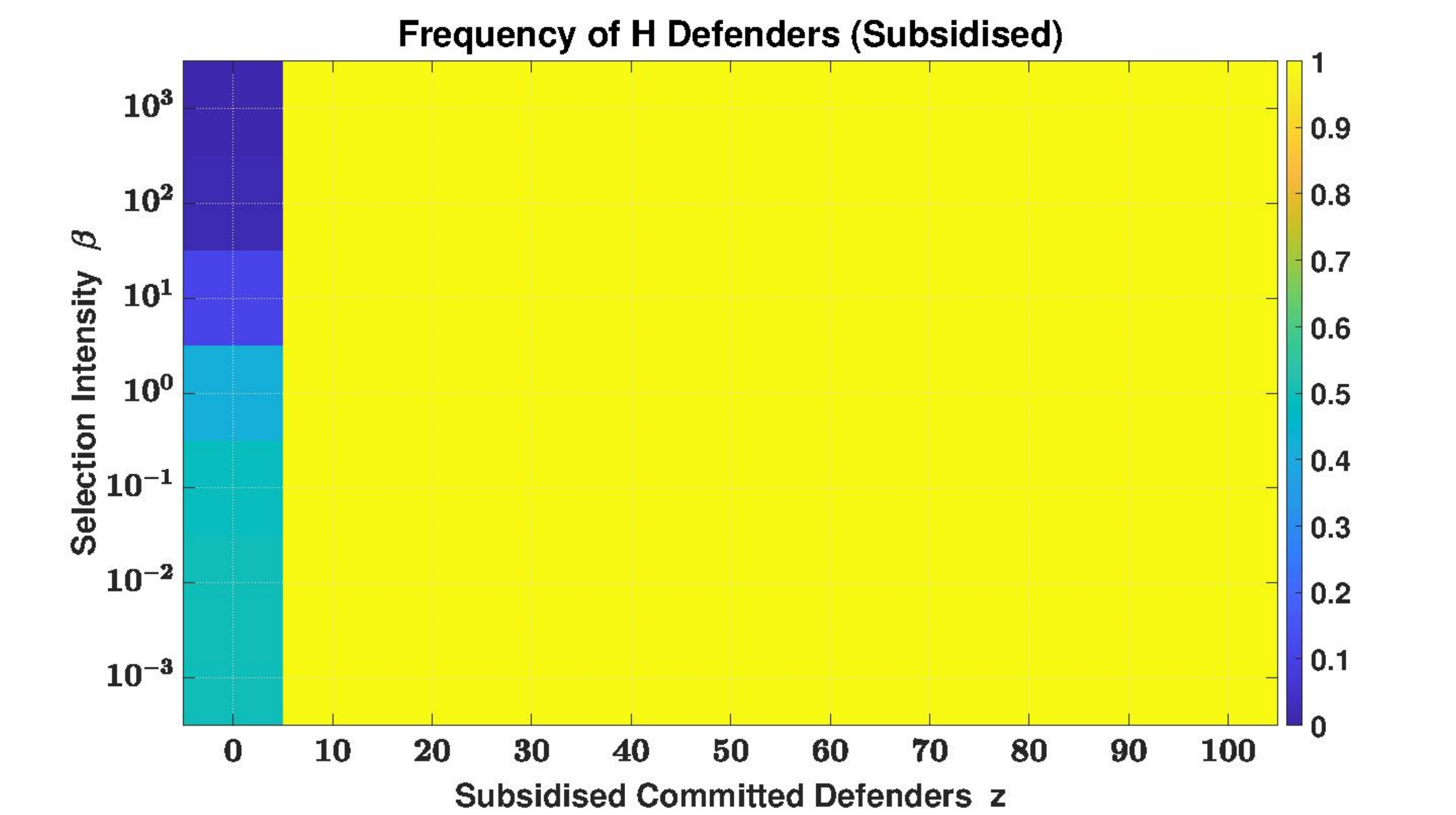}
    \caption{}
  \end{subfigure}
  \caption{\textbf{Stationary outcomes under subsidised committed defenders.}
Heatmaps show that subsidy reduces the effective cost of high defence, increasing the payoff advantage of $H$ defenders. As a result, the system rapidly shifts toward the secure equilibrium $(NA,H)$, which becomes stable at low levels of commitment (approximately $z \approx 6$) enabling efficient and robust adoption of high defence across a wide range of $\beta$. Baseline parameters are $N=100$, $c_{ah}=0.85$, $b_{ah}=1.90$, $c_{al}=0.10$, $b_{al}=1.60$, $p_{dH}=0.82$, $p_{dL}=0.75$, $B_H=0.75$, $B_L=0.55$, $C_H=0.41$, $C_L=0.20$, $W_H=0.22$, and $W_L=0.10$.}
\label{fig:16}
\end{figure}

\end{document}